\pdfoutput=1

\documentclass[11pt]{article}

\usepackage[final]{acl}

\usepackage{times}
\usepackage{latexsym}
\usepackage{times}
\usepackage{booktabs}
\usepackage{latexsym}
\usepackage{tabularray}
\usepackage{rotating}
\usepackage{bm}
\usepackage{amssymb}
\usepackage{colortbl}
\usepackage{xcolor}
\usepackage{longtable}

\usepackage{pifont}

\usepackage[rgb]{xcolor}

\selectcolormodel{natural}
\usepackage{ninecolors}
\selectcolormodel{rgb}
\usepackage{tabularray}

\usepackage{caption}
\usepackage[most]{tcolorbox}
\usepackage{hyperref}
\usepackage[normalem]{ulem}
\usepackage{makecell}
\usepackage{caption}
\usepackage{multirow}
\usepackage{rotating}
\usepackage[most]{tcolorbox}
\usepackage{hyperref}
\usepackage[normalem]{ulem}
\usepackage{booktabs}
\newcommand\hlwrong{\bgroup\markoverwith
  {\textcolor[RGB]{255, 189, 189}{\rule[-.5ex]{2pt}{2.5ex}}}\ULon}

\usepackage{tikz}
\usepackage{subcaption}
\usepackage{supertabular}
\usepackage{collcell}
\usepackage{makecell}
\usepackage{multirow}
\usepackage{array}
\usepackage{graphicx}
\usepackage{subcaption}

\definecolor{lightgrey}{RGB}{235, 236, 237}
\definecolor{darkgrey}{RGB}{124, 124, 125}

\definecolor{lightteal}{RGB}{208,223,226}
\definecolor{teal}{RGB}{69,129,129}

\definecolor{lightorange}{RGB}{252,229,205}
\definecolor{burntorange}{RGB}{207,146,82}

\definecolor{lightpurple}{RGB}{217,210,233}
\definecolor{darkpurple}{RGB}{124,102,179}
\definecolor{coolgreen}{RGB}{73, 176, 104}
\definecolor{coolred}{RGB}{235, 125, 120}

\usepackage[T1]{fontenc}

\usepackage[utf8]{inputenc}

\usepackage{microtype}

\usepackage{inconsolata}
\usepackage{hyperref}

\hypersetup{colorlinks=true,linkcolor=blue,urlcolor=blue}

\usepackage{tikz}
\definecolor{lightteal}{RGB}{234,209,220}
\definecolor{teal}{RGB}{116,27,71}
\definecolor{lightblue}{RGB}{181, 179, 242}
\definecolor{darkblue}{RGB}{68, 63, 204}

\definecolor{darkred}{RGB}{150,30,45}

\newcommand{\abr}[1]{\textsc{#1}}
\newcommand{\unite}{\abr{Unite-Dev}}
\newcommand{\bird}{\abr{Bird-Dev}}
\newcommand{\spider}{\abr{Spider-Dev}}
\newcommand{\realistic}{\abr{Spider-Realistic}}
\newcommand{\A}{$\mathcal{D}_A$}
\newcommand{\target}{$\mathcal{D}_\text{target}$}
\newcommand{\B}{$\mathcal{D}_B$}
\newcommand{\sqlspace}{\texttt{SQLSpace}}

\newenvironment{tight_enumerate}{
\begin{enumerate}
  \setlength{\itemsep}{0pt}
  \setlength{\parskip}{0pt}
}{\end{enumerate}}

\newcommand{\checkmarksmall}{\includegraphics[height=10pt,width=10pt]{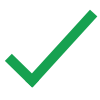}}
\newcommand{\crossmarksmall}{\includegraphics[height=10pt,width=10pt]{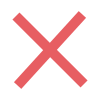}}

\newtcolorbox[list inside=prompt,auto counter,number within=section]{prompt}[1][]{
    colbacktitle=black!60,
    fonttitle=\small,
    coltitle=white,
    fontupper=\footnotesize,
    boxsep=4pt,
    left=0pt,
    right=0pt,
    top=0pt,
    bottom=0pt,
    boxrule=1pt,
    #1,
}

\title{\texttt{SQLSpace}: A Representation Space for Text-to-SQL to \\Discover and Mitigate Robustness Gaps}

\newcommand{\aspace}{\hspace{1em}}
\newcommand{\umd}{$^{\spadesuit}$}
\newcommand{\adobe}{$^{\clubsuit}$}

\author{%
    \textbf{Neha Srikanth}\umd \thanks{~\,Work on the \sqlspace~representation was completed during an internship at Adobe Research. Analysis of model performance for open-source models was done at the University of Maryland after the completion of the internship.}\aspace
    \textbf{Victor Bursztyn}\adobe \aspace
    \textbf{Puneet Mathur}\adobe \aspace
    \textbf{Ani Nenkova}\adobe \aspace\\
    \umd{}University of Maryland\aspace \adobe{}Adobe Research \\
    \texttt{nehasrik@umd.edu}
}

\begin{document}
\maketitle
\begin{abstract}
We introduce \texttt{SQLSpace}, a human-interpretable, generalizable, compact representation for text-to-SQL examples derived with minimal human intervention.
We demonstrate the utility of these representations in evaluation with three use cases: \textit{(i)} closely comparing and contrasting the composition of popular text-to-SQL benchmarks to identify unique dimensions of examples they evaluate, 
\textit{(ii)} understanding model performance at a granular level beyond overall accuracy scores, and \textit{(iii)} improving model performance through targeted query rewriting based on learned correctness estimation.
We show that \sqlspace~enables analysis that would be difficult with raw examples alone: it reveals compositional differences between benchmarks, exposes performance patterns obscured by accuracy alone, and supports modeling of query success.

\end{abstract}

\section{Introduction}
Systems tasked with translating a natural language utterance to an executable SQL query (text-to-SQL, or NL2SQL for short) are typically evaluated for their accuracy on one or more benchmarks such as \abr{Bird}~\cite{li2024bird} or \abr{Spider}~\cite{lei2024spider}. 
While these accuracies can be used to rank models on a leaderboard in a coarse-grained way, this evaluation paradigm obscures parts of evaluation that are useful for researchers and practitioners to know.
It cannot answer naturally arising questions around dataset composition (\textit{Why are the accuracies for the same model so different on the two benchmarks?}), open challenges for the field that future work must address (\textit{What subsets of data are easy or difficult for all models?}), and explorations into performance and cost trade-offs (\textit{Are there subsets of data on which cheaper models perform as well as more expensive models?}).

\begin{figure}[t!]
\centering
\includegraphics[scale=0.55]{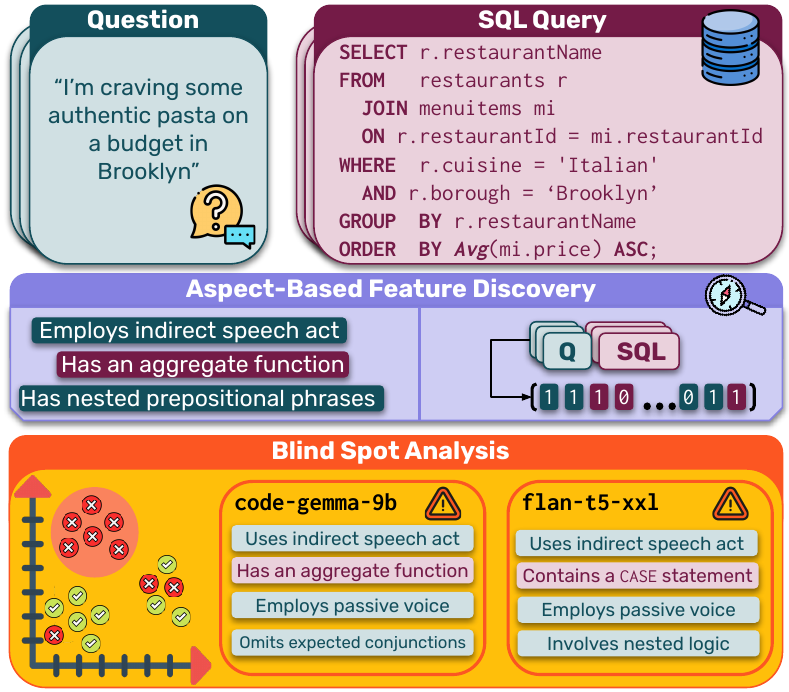}
\caption{Our framework generates compact representations of NL2SQL examples by ingesting a dataset, discovering shared properties of dataset items in natural language, and evaluating these properties on examples to produce binary feature vectors. Clustering these feature vectors and examining a model's cluster-level accuracy reveals classes of examples that it systematically struggles with, called \textit{blind spots}.}
\label{fig:teaser-example}
\end{figure}

A deeper, finer-grained understanding of text-to-SQL examples could help us better understand model performance and benchmark composition, making it easy for practitioners to compare their own data to existing benchmarks in order to select the top-performing model on a benchmark that best reflects their own data. 
Similarly, such understanding would help researchers looking to build new robustness benchmarks better analyze existing gaps in datasets and systems in order to inform the design of new challenge evaluation sets.

To facilitate informed decision-making in scenarios like these, we introduce \texttt{SQLSpace}, a framework to increase visibility into benchmark composition and model behavior with minimal human intervention.\footnote{\url{https://github.com/nehasrikn/robust-sql}.}
\sqlspace~ingests one or more NL2SQL benchmarks and extracts human-interpretable features of examples (e.g. \textit{``has an aggregate function''}) in order to construct generalized vector representations of any NL2SQL example (\S\ref{sec:methodology}) in a largely automated manner.

We demonstrate the utility of these vector representations with three use cases that leverage the representations in different ways.
First, we compare the distribution of examples in popular NL2SQL benchmarks (\S\ref{sec:benchmarks}), allowing us to understand which particular dimensions each dataset uniquely evaluates.
Then, we compare the strengths and weaknesses of 13 models with greater detail than in leaderboards by clustering examples across benchmarks that share similar features, and identifying ``blind spots'', or systematic classes of examples that a model struggles with (\S\ref{sec:blind-spots}).
We identify two clusters of universal blind spots and multiple clusters for which cheaper models can outperform expensive ones.
Finally, show that learning a correctness estimator for a given model to inform rewrites of NL queries that are predicted to fail can improve accuracy (\S\ref{sec:rewriting}).

\section{Background}
\label{sec:background}
\paragraph{Task and Evaluation.} Text-to-SQL involves translating natural language question $X$ to a valid SQL query $Y$ that is executable over a database with schema $\mathcal{S}$.
Along with knowledge of $\mathcal{S}$ and SQL, this task may require other domain knowledge or forms of linguistic and inferential reasoning~\cite{gan-etal-2021-spider-dk}.
Generated queries $Y$ are typically evaluated by executing them and comparing the resulting output with that produced by executing the gold query~\cite{li2024bird}, and computing execution accuracy (EX), or the proportion of examples for which the results of the predicted and gold SQL queries match.
While other metrics, such as PCM-F1~\cite{hazoom-etal-2021-text}, have been proposed to award partial credit, we report EX as our primary evaluation metric due to its wider adoption in well-established benchmarks \cite{li2024bird, lei2024spider}.

\paragraph{Datasets.} 
Datasets such as \abr{Spider}~\cite{yu-etal-2018-spider, lei2024spider} and \abr{Bird-Bench}~\cite{li2024bird} are general-purpose benchmarks designed to reflect realistic queries and use-cases, spanning several database schemas and domains.
These benchmarks maintain leaderboards to facilitate efficient and standardized evaluation and are helpful in gauging model performance.
While some benchmarks release accompanying analyses on the composition of examples (e.g. \citet{yu-etal-2018-spider} include SQL pattern coverage, or ``difficulty'' metadata), these statistics are not standardized across datasets (Appendix~\ref{appendix:bird-vs-spider-vs-sqlspace}). 
In turn, it is challenging to identify qualitative differences between benchmarks and, more importantly, understand the specific strengths and weaknesses of the models evaluated on them.

\textit{Robustness}-oriented benchmarks help address this by targeting specific model weaknesses.
\citet{gan-etal-2021-spider-syn} create \abr{Spider-Syn} to measure paraphrastic robustness~\cite{srikanth-etal-2024-often} after finding models vulnerable to synonym substitution~\cite{utama2018end, ma2021mt}.
Other \abr{Spider} variants incorporate domain knowledge~\cite{gan-etal-2021-spider-dk}.
\citet{pi-etal-2022-table} study model robustness to table perturbations, and \citet{chang2023dr} explore 17 perturbation types to both questions and SQL queries.
These types of robustness benchmarks typically rely on hand-designed perturbations informed by human priors, requiring significant manual effort.
Experts must identify systematic model failures and craft challenge sets targeting those errors, as in \citet{chang2023dr}.
As models improve, this process resembles an iterative cycle where researchers (1) identify remaining model weaknesses, (2) design challenge sets to target those weaknesses, after which (3) models are optimized to saturate those challenge benchmarks.
Not only is this process costly, but it risks overlooking robustness gaps for particular \textit{combinations} of properties of NL2SQL examples, or more broadly, along other dimensions not explicitly included in these benchmarks.
\sqlspace~addresses these gaps in NL2SQL model robustness evaluation with minimal human intervention (Appendix~\ref{appendix:human-intervention}).

\section{Representation Discovery}
\label{sec:methodology}
Understanding the composition of NL2SQL datasets offers both theoretical and practical benefits.
It facilitates a deeper scientific understanding of the NL2SQL task by identifying requisite reasoning abilities as well as informed decision-making about model and benchmark selection for different use cases.
We introduce a method, \texttt{SQLSpace}, to discover interpretable feature-based representations of NL2SQL examples which we use for downstream analysis on benchmarks (\S\ref{sec:benchmarks}) and model performance (\S\ref{sec:blind-spots}).

\subsection{Motivation}
\label{subsec:motivation}
\begin{figure}
\centering
\includegraphics[scale=0.78]{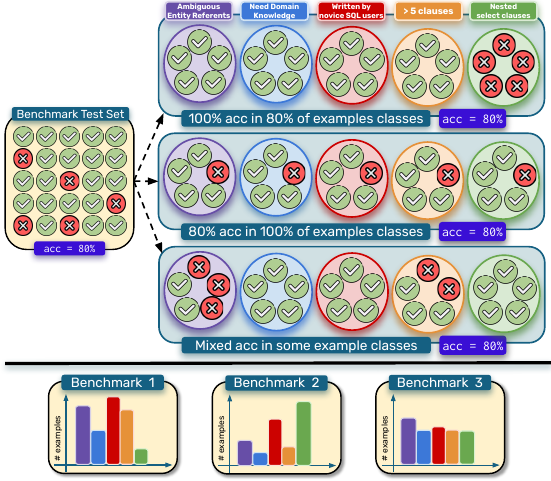}
\caption{Aggregate accuracy may obscure important performance characteristics of models. Three models with identical 80\% accuracy on benchmark $B$ exhibit different error patterns across example classes, while varying class distributions across benchmarks can explain performance differences.}
\label{fig:motivation}
\end{figure}

We motivate \sqlspace~with a simplified thought experiment designed to illustrate the opacity of accuracy-based leaderboard evaluation (Figure~\ref{fig:motivation}).
Consider a model $M$ that scores 80\% accuracy on benchmark $B$. Conventional NL2SQL leaderboards simply rank models by this accuracy, obscuring deeper performance characteristics of $M$. 
However, grouping examples according to shared properties (e.g., queries containing nested \texttt{SELECT} clauses or ambiguous entity references) and analyzing model predictions within and across these groups could reveal more about the strengths and weaknesses of $M$.
For this thought experiment, we assume exclusive class membership, though we revisit this in \S\ref{sec:blind-spots}.

Fig.~\ref{fig:motivation} illustrates three scenarios that all maintain the 80\% accuracy of $M$ while exhibiting fundamentally different error distributions: (1) perfect accuracy on certain classes and complete failure on others, (2) uniform performance across all classes, and (3) mixed performance within individual classes.
This breakdown helps us better understand $M$.
Furthermore, the lower panel of Fig.~\ref{fig:motivation} illustrates that example class distributions may vary significantly across benchmarks (\S\ref{sec:benchmarks}).
If benchmark $B$ contains disproportionately many examples with nested clauses relative to other evaluation sets, this compositional bias could account for $M$'s degraded performance on $B$, since the upper panel shows a scenario in which $M$ systematically fails on those types of examples.
The \sqlspace~representations we build in this section help us automatically identify such shared properties, facilitating scalable benchmark and model analysis.

\subsection{Representation Construction}
\sqlspace~ingests a set of examples and involves four steps to semi-automatically produce general-purpose vector representations of examples: (1) aspect-based example \textbf{description generation}, (2) \textbf{feature discovery from descriptions}, (3) \textbf{feature deduplication}, and lastly, (4) \textbf{representation construction (inference) on examples}.

\paragraph{Development Example Set.} We use a portion of the development set from the \abr{Unite} corpus~\cite{lan2023unite} (\unite) for automatic feature discovery since it collates several public NL2SQL datasets.
The portion we use includes a total of 10,697 examples from \abr{Spider}~\cite{yu-etal-2018-spider}, \abr{Squall}~\cite{shi-etal-2020-squall}, \abr{Spider-Syn}~\cite{gan-etal-2021-spider-syn}, \abr{Criteria2SQL}~\cite{yu2020criteria2sql}, \abr{SparC}~\cite{yu-etal-2019-sparc}, \abr{CoSQL}~\cite{yu-etal-2019-cosql}, \abr{Spider-DK}~\cite{gan-etal-2021-spider-dk}, \abr{ParaphraseBench}~\cite{utama2018end}, \abr{KaggleDBQA}~\cite{lee-etal-2021-kaggledbqa}, \abr{ACL-SQL}~\cite{kaoshik2021aclsql}, \abr{SEOSS-Queries}~\cite{tomova2022seoss}, and \abr{FIBEN}~\cite{sen2020fiben}. 
Examples in these benchmarks span various natural language constructions, SQL patterns, and database schemas, making the compilation well-suited for discovering generalizable features.
Though we select \unite, our method is dataset-agnostic, and in practice can be applied to any example collection. 

\begin{table}[t!]
\centering
\tiny
\resizebox{\columnwidth}{!}{
\begin{tabular}{lclc} 
\toprule
\multicolumn{4}{p{8cm}}{\colorbox{lightorange}{\textcolor{burntorange}{\textbf{Schema ($\mathcal{S}$):}}} \texttt{\textbf{employee}: id, name, salary, dept\_id | \textbf{department}: dept\_id, name}} \\[1ex]
\multicolumn{4}{p{8cm}}{\colorbox{lightpurple}{\textcolor{darkpurple}{\textbf{Question ($X$):}}} \textit{How many employees earn more than their department's average salary?}} \\[1ex]
\multicolumn{4}{p{8cm}}{\colorbox{lightteal}{\textcolor{teal}{\textbf{SQL ($Y$):}}} \texttt{SELECT COUNT(*) FROM employee e WHERE salary > (SELECT AVG(salary) FROM employee WHERE dept\_id = e.dept\_id)}} \\[1ex]
\toprule
\textbf{Aspect} & \textbf{\#} & \textbf{Example Predicates ($p$)} & \textbf{Evaluated} \\
\midrule
\textbf{Syntax} & 74 & omits expected conjunctions & 0 \\
 & (40\%) & contains subordinate clauses & 1 \\
 &  & contains nested conditionals & 0 \\ 
\midrule
\textbf{SQL} & 41 & includes a subquery & 1 \\
\textbf{Syntax} & (22\%) & has an aggregate function & 1 \\
 &  & contains a \texttt{CASE} statement & 0 \\
\midrule
\textbf{Example} & 27 & involves nested logic & 1 \\
\textbf{Semantics} & (14\%) & has direct relationship & 0 \\
 &  & requires domain knowledge & 0 \\ 
\midrule
\textbf{Pragmatics} & 33 & employs direct speech acts & 1 \\
 & (18\%) & relies on conversational impl. & 0 \\
 &  & exhibits minimal ambiguity & 1 \\ 
\midrule
\textbf{Database} & 17 & uses exact column names & 1 \\
\textbf{Reasoning} & (9\%) & requires commonsense reasoning & 0 \\
 &  & mirrors schema structure & 1 \\
\midrule
\textbf{Total} ($\mathcal{P}$) & \textbf{187} (100\%) & & \\
\bottomrule
\end{tabular}}
\caption{Our final set of natural language predicates, $\mathcal{P}$, spans multiple aspects of NL2SQL examples. Building a representation of an example $e$ involves evaluating each predicate $p$ on $e$ to produce binary feature vector.}
\label{tab:example-predicates}
\end{table}

\paragraph{Step 1: Description Generation.} 
Building a human-interpretable example representation requires identifying \textit{diverse} and \textit{meaningful} properties of NL2SQL examples that may not necessarily be captured by traditional embedding methods.\footnote{We experimented with instruction-tuned embedding models for generating representations, but found they focused on shallow features, even after masking schema-specific entities.}

\begin{figure*}
\centering
\includegraphics[scale=0.53]{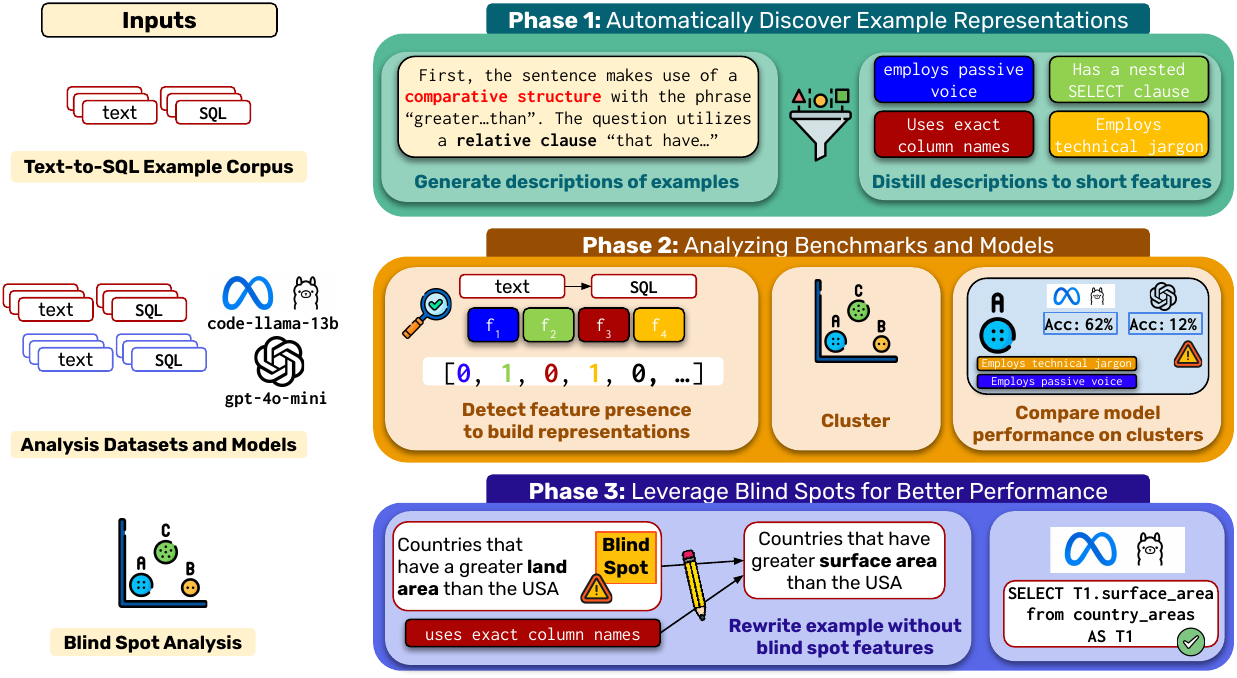}
\caption{We discover representations of NL2SQL examples, and then use these representations in two applications: fine-grained analysis of models and benchmarks, and improving performance of models by rewriting NL questions to eliminate features associated with incorrect predictions.}
\label{fig:pipeline}
\end{figure*}

For an example $e$ with natural language question $X$ and gold SQL query $Y$, a \textbf{\texttt{Describer}} model generates five descriptions of $e$ that focus on different ``aspects'' related to the NL2SQL task:

\begin{tight_enumerate}
  \item \textbf{Syntax} ($d_\text{syn})$: Commentary on the linguistic syntax of $X$, including word order, grammatical relations, sentence structure, etc.
  \item \textbf{SQL Syntax} ($d_\text{sql-syn}$): Covers elements of $Y$, including the structure and complexity of the query as well as SQL keywords or other syntactic elements.
  \item \textbf{Example Semantics} ($d_\text{sem}$): Specific commentary on the relationship between $X$ and $Y$, such as parallel characteristics, types of reasoning required to map between $X$ and $Y$, and other similarities and differences.
  \item \textbf{Pragmatics} ($d_\text{prag}$): Discussion of pragmatic elements of $X$ such as speech acts~\cite{searle1975taxonomy}, adherence to Gricean maxims~\cite{grice1975logic}, uses of presupposition~\cite{beaver1997presupposition} and implicature~\cite{grice1975logic}, and relevance. May also include commentary on any ambiguity~\cite{wang-etal-2023-ambigSQL} or vagueness~\cite{saparina2024ambrosia}.
  \item \textbf{Database Reasoning} ($d_\text{db}$): Commentary on the relationship between $X$ and the database schema $\mathcal{S}$, including necessary reasoning required to map between entities in $X$ and columns in  $\mathcal{S}$, etc.
\end{tight_enumerate}

\noindent Inspired by aspect-based summarization~\cite{angelidis-lapata-2018-aspect}, these descriptions are designed to capture fine-grained details about examples in the input dataset.
The choice of aspects above is informed by previous studies on the types of reasoning required to solve NL2SQL problems.
We select \texttt{gpt-4o-2024-05-13} as our \texttt{Describer}\footnote{\sqlspace~can be run using any \texttt{Describer}, open-source or proprietary. Future work may explore the effects of different \texttt{Describers} on generated features.} and generate these five aspect-based descriptions for all 10,697 examples in \unite~using Prompts~\ref{prompt:describe-syntax}--\ref{prompt:describe-db} (see Table~\ref{tab:description-examples} for example descriptions).

\paragraph{Step 2: Feature Discovery.}
Long-form descriptions generated by the \texttt{Describer} in Step 1 serve as a repository of aspect-related facts about an example.
Examples similar in certain aspects may share phrases or sentences in their descriptions.
To discover shared properties across descriptions, we use a module from the goal-driven explainable clustering pipeline in \citet{wang-etal-2023-goal} designed to propose binary \textit{natural language predicates} from descriptions. These predicates then serve as candidate features of an NL2SQL example (henceforth, we use \textit{predicate} and \textit{feature} interchangeably).
Given a natural language predicate $p$ and a NL2SQL example $e$, $p(e) = 1$ when $e$ expresses $p$, and $p(e) = 0$ when $e$ does not express $p$.

We run a predicate proposal module from \citet{wang-etal-2023-goal} which ingests a collection of texts and a natural language goal, and has a \textbf{\texttt{Proposer}} LLM generate a list of candidate predicates, essentially performing ``in-context clustering.''
Concretely, this process iteratively (1) samples a random subset of documents from the input collection and (2) prompts the \texttt{Proposer} to generate a structured list of $n$ predicates, repeating the process for $j$ iterations.

For each collection of aspect-based descriptions $\mathcal{C_\text{aspect}}$ (e.g. $\mathcal{C}_{syn} = \{d_{syn_1}, d_{syn_2}, ...\}$), the \texttt{Proposer} generates $n=40$ predicates per iteration for $j=5$ iterations (Appendix~\ref{appendix:predicate-discovery}).
We use \texttt{gpt-3.5-turbo-0125} as our \texttt{Proposer} (see Table~\ref{tab:example-predicates} for example predicates).
\textbf{Note that describing examples in Step 1 before predicate proposal disentangles reasoning and understanding examples from the task of finding commonalities across examples in this step.}
We also use \texttt{gpt-4o-2024-08-06} as an additional \texttt{Proposer} to diversify candidate predicates, as different models prioritize different parts of descriptions when proposing predicates (see Table~\ref{tab:proposed-counts} for counts of proposed candidate predicates).

\paragraph{Step 3: Feature Deduplication.}
The pool of candidate predicates for each aspect generated by the \texttt{Proposers} in the previous step contains duplicates (\textit{``contains a nested \texttt{JOIN}''} and \textit{``uses nested \texttt{JOINS}''}).
We remove them by filtering out those with high token similarity as measured by Levenshtein distance,\footnote{We use \href{https://github.com/seatgeek/thefuzz}{\texttt{thefuzz}} with a token set similarity threshold of $\epsilon=70$. This method, based on the Levenshtein edit distance, compares sets of tokens, ignoring order and redundancies. It effectively removes predicates with similar wording (e.g., \textit{contains a \texttt{JOIN}''} and \textit{uses a \texttt{JOIN}''}).} and then manually removing any remaining paraphrases to ensure a clean set.\footnote{We also experimented with automatic methods such as instruction-tuned embedding models, but ultimately relied on manual deduplication to ensure a clean final predicate set.}
This process yields 187 general-purpose features across five aspects (Table~\ref{tab:example-predicates}), henceforth denoted as $\mathcal{P}$, which we exhaustively list in Table~\ref{tab:full-list} as an artifact of our work that other researchers may leverage.

\paragraph{Step 4: Example Representation Construction.}
We now have a set of predicates that can serve as descriptive features of NL2SQL examples and that span all components of the example: the natural language question $X$, its relationship to the schema $\mathcal{S}$, and the corresponding SQL query $Y$. 
We create a vector representation of an example $e$ by eliciting binary judgments from a predicate \textbf{\texttt{Evaluator}} with Prompts~\ref{prompt:pred-eval-syntax}--\ref{prompt:pred-eval-db} on $\{p(e) \mid p \in \mathcal{P}\}$~\cite{wang-etal-2023-goal}, yielding a $|\mathcal{P}|$-dimensional binary vector. 
We use \texttt{gpt-4o-2024-08-05} as our \texttt{Evaluator}.
For each aspect, we only include the relevant components of the example in the predicate evaluation prompt (e.g., syntax-based predicates are only evaluated over the natural language question, or database reasoning-predicates are only evaluated over the natural language question and the schema). 

We estimate the accuracy of our chosen \texttt{Evaluator} by randomly sampling 50 example-predicate pairs for each aspect (250 examples total) and have two authors independently evaluate the predicate on the example.
The average accuracy of the \texttt{Evaluator} based on the two authors was 73\%.
We compute the agreement between the two annotators, obtaining Cohen's kappa values of $\kappa=60.2$, indicating substantial agreement~\cite{artstein2008inter} (see Table~\ref{tab:predicate-eval} for statistics per aspect).
While we use a larger closed-source model to evaluate predicates for proof-of-concept, future work could explore using fine-tuning lighter-weight models on the task of predicate evaluation. 

\paragraph{What do these example representations help achieve?}
Our four-step pipeline (see Appendix~\ref{sec:ablation} for ablation discussion and further intuitions) produces $|\mathcal{P}|$-dimensional binary vectors that serve as a \textbf{general-purpose unified representation for any NL2SQL example}.
\begin{figure*}[ht!]
\centering
\includegraphics[scale=0.5]{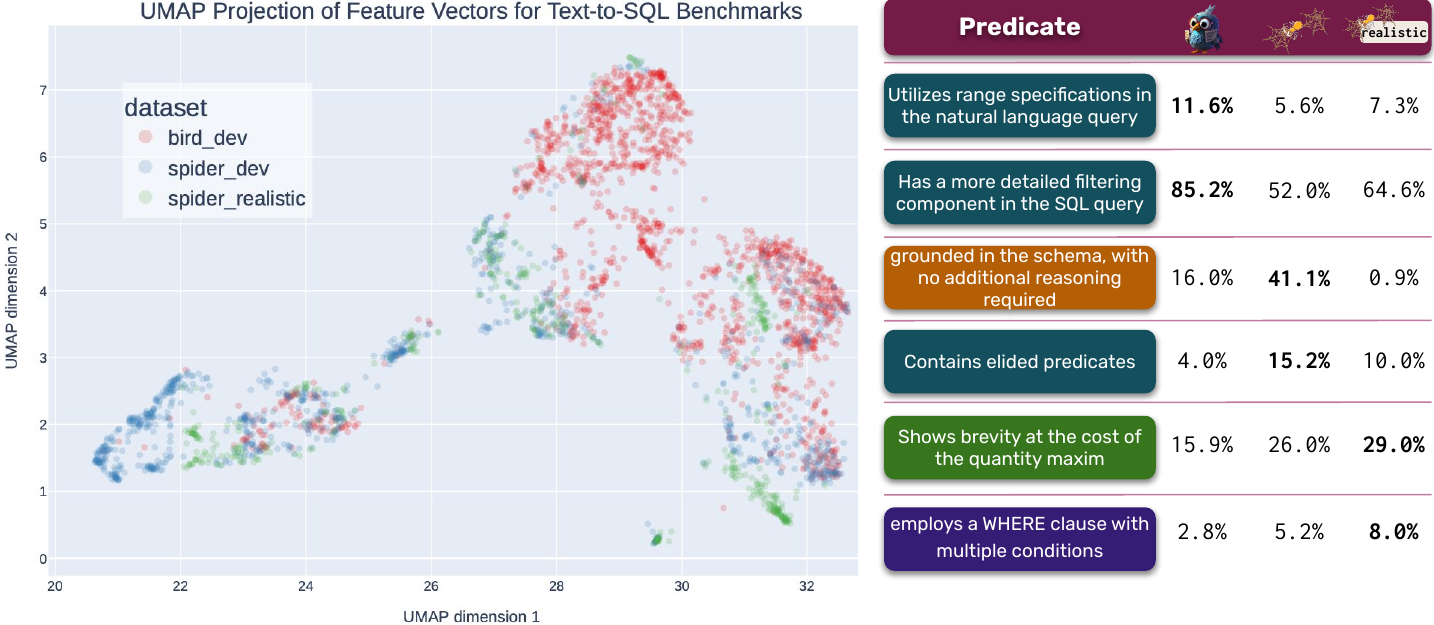}
\caption{Visualizing a \abr{UMAP} projection (left) of our example representations for three NL2SQL datasets reveals classes of examples across datasets that share certain properties. Computing the proportions of examples exhibiting certain features (right) reveals dimensions along which the composition of datasets statistically significantly differs.}
\label{fig:umap-corpora}
\end{figure*}
These representations enable various analyses, including comparing dimensions along which benchmarks significantly differ (\S\ref{sec:benchmarks}) as well as understanding fine-grained classes of examples across benchmarks on which NL2SQL models struggle (\S\ref{sec:blind-spots}).
They also allow us to build a correctness classifier to estimate the likelihood that a model will produce correct SQL for a natural language example, and in cases where an example exhibits features associated with blind spots, intervene to remove them (\S\ref{sec:rewriting}).

\section{Comparing Text-to-SQL Benchmarks}
\label{sec:benchmarks}
Analyzing the composition of datasets can reveal distributions of example properties \textit{within} datasets, and highlight similarities and differences of examples \textit{across} benchmarks.
These analyses enhance our understanding of the reasoning skills or knowledge needed to perform well on NL2SQL, and in turn, can help inform decisions about model design or new benchmark construction.
For instance, discovering the prevalent syntactic complexity of questions may inspire new linguistically-informed approaches to modeling or data creation.

Dataset comparison also offers practical benefits.
When a practitioner has a proprietary NL2SQL dataset, \target, understanding prevalent properties of examples in \target~as well as the dimensions along which it resembles or differs from existing benchmarks \A~or \B~helps them select the cheapest model that performs well on their data, and opens up other avenues such as data augmentation with benchmarks that closely resemble \target.

\paragraph{Setup.} We compute $|\mathcal{P}|$-dimensional vector representations for examples (Step 4 in \S\ref{sec:methodology}) in three different datasets: (1) the development set (1,034 examples) of \abr{Spider}~\cite{yu-etal-2018-spider}, one of the prevailing cross-domain benchmarks in the NL2SQL community (\spider), (2) the development set (1,534 examples) of \abr{Bird-Bench}~\cite{li2024bird}, another popular large-scale cross-domain benchmark released as an updated, more challenging alternative to \spider (\bird), and (3) \realistic~\cite{deng-etal-2021-structure}, a dataset of 508 examples based off of \spider~designed to reflect more realistic natural language questions.
We pose a scenario in which \realistic~is a target dataset whose composition we want to better understand as compared to established benchmarks \spider~(\A) and \bird~(\B) to illustrate the utility of our vector representations. 

\paragraph{Comparing \spider~and~\bird.}Figure~\ref{fig:umap-corpora} visualizes a \abr{UMAP} projection~\cite{mcinnes2018umap} of feature vectors of all examples in \A, \B, and \target~into two dimensions.\footnote{To project feature vectors using \abr{UMAP}, we use a neighbor count (\texttt{n\_neighbors}) of 50 and a minimum distance (\texttt{min\_dist}) of 0.01.}
While dataset-specific clusters do emerge for \spider~and \bird, Figure~\ref{fig:umap-corpora} reveals classes of examples that may share similar properties in $\mathcal{P}$.
Referencing hand-annotated difficulty metadata released for each example in \bird~reveals that these high regions of overlap between \bird~and \spider~mainly occur with \bird~examples annotated as ``simple'' (Figure~\ref{fig:bird-v-spider}).
This aligns with our priors on the construction of both datasets, as \bird~was introduced as a more difficult benchmark due to its complex schema~\cite{li2024bird}, and features that are often associated with \textit{simpler} examples (use of exact column names in the natural language question, or clear mappings between the question and the SQL query) are more prevalent in \spider~examples~\cite{deng-etal-2021-structure}.

\paragraph{\target.} We observe regions with substantial overlap between \target~and our two benchmarks, along with a couple of outlier clusters of \target-specific examples.
To interpret these patterns, we compute the proportion of examples in each dataset exhibiting each feature and run a chi-square test to identify features with statistically significant differing proportions.
This allows us to understand the dimensions along which each dataset is distinct.

The table on the right in Figure~\ref{fig:umap-corpora} shows six illustrative features and the proportion of examples in each dataset that express them.
The third row shows a question-schema alignment feature rarely present in examples from \realistic~(\target).
This validates our approach, since it aligns with the creation of \realistic, which was constructed by \citet{deng-etal-2021-structure} by manually modifying natural language questions in \spider~to remove or paraphrase mentions of column names.
A likely consequence of this process appears in Row 5, where \realistic~contains the highest proportion of examples that trade satisfying the Gricean maxim of quantity~\cite{grice1975logic} for brevity.

These analyses allow us to better understand the skills required to perform well on each dataset (e.g., \bird~contains more syntactically complex examples than \realistic: 20\% of examples include mixed use of symbols and words as compared to 5\% in \spider~and 6\% in \realistic).
They illustrate how our framework can equip a user with detailed knowledge about their dataset when they are trying to make sense of new \target~data, and can be repeated with any other dataset.

\section{Identifying Model Blind Spots}
\label{sec:blind-spots}
\newcommand{\cluster}[1]{$\mathbf{C_{#1}}$}

\definecolor{birdcolor}{RGB}{86,180,233}
\definecolor{spidercolor}{RGB}{230,159,0}
\definecolor{realisticcolor}{RGB}{0,158,115}

\definecolor{lightblue}{RGB}{235, 245, 255}
\definecolor{mediumblue}{RGB}{200, 220, 255}

\newcommand{\coloredaccuracy}[2]{%
    \cellcolor{blue!#1!white!40}%
    $#1^{~(#2)}$%
}

\newcommand{\minipie}[2]{%
    \begin{tikzpicture}[scale=0.4]
    \pgfmathsetmacro{\angleone}{#1*3.6}
    \pgfmathsetmacro{\angletwo}{#2*3.6}
    
    \pgfmathsetmacro{\cumone}{\angleone}
    
    \draw[fill=birdcolor] (0,0) -- (0:1) arc (0:\angleone:1) -- cycle;
    \draw[fill=spidercolor] (0,0) -- (\angleone:1) arc (\angleone:360:1) -- cycle;
    \end{tikzpicture}%
}

\newcommand{\modelscores}[3]{%
    \textbf{\texttt{#1}} ({\small\textcolor{birdcolor}{$#2$}\,/\,\textcolor{spidercolor}{$#3$}})%
}

\begin{table*}
\centering
\resizebox{2.08\columnwidth}{!}{
\begin{tabular}{lcccccccccccccc}
\toprule
\textbf{Cluster Number} & $\mathbf{C_1}$ & $\mathbf{C_2}$ & $\mathbf{C_3}$ & $\mathbf{C_4}$ & $\mathbf{C_5}$ & $\mathbf{C_6}$ & $\mathbf{C_7}$ & $\mathbf{C_8}$ & $\mathbf{C_9}$ & $\mathbf{C_{10}}$ & $\mathbf{C_{11}}$ & $\mathbf{C_{12}}$ & $\mathbf{C_{13}}$ & $\mathbf{C_{14}}$ \\
\midrule
\# total examples & $148$ & $303$ & $194$ & $203$ & $87$ & $146$ & $234$ & $230$ & $275$ & $160$ & $147$ & $130$ & $161$ & $150$ \\
\midrule
Distribution & \minipie{35.1}{64.9} & \minipie{83.8}{16.2} & \minipie{86.1}{13.9} & \minipie{45.3}{54.7} & \minipie{14.9}{85.1} & \minipie{69.2}{30.8} & \minipie{6}{94} & \minipie{97}{3} & \minipie{93.8}{6.2} & \minipie{83.1}{17} & \minipie{27.9}{72.1} & \minipie{3.1}{96.9} & \minipie{72}{28} & \minipie{44}{56} \\
\tikz\draw[fill=birdcolor] (0,0) rectangle (0.3,0.3);~\bird~($1534$) & $52$ & $254$ & $167$ & $92$ & $13$ & $101$ & $14$ & $223$ & $258$ & $133$ & $41$ & $4$ & $116$ & $66$ \\
\tikz\draw[fill=spidercolor] (0,0) rectangle (0.3,0.3);~\spider~($1034$) & $96$ & $49$ & $27$ & $111$ & $74$ & $45$ & $220$ & $7$ & $17$ & $27$ & $106$ & $126$ & $45$ & $84$ \\
\hline
\%\realistic~($508$) & $6.7$ & $8.9$ & $4.1$ & $9.1$ & $4.9$ & $16.5$ & $11.6$ & $3.7$ & $3.1$ & $2.8$ & $12.4$ & $2.6$ & $3.7$ & $9.8$ \\
\midrule
\modelscores{gpt-4o-mini}{37.1}{76.4} & \coloredaccuracy{84.5}{13} & \coloredaccuracy{56.1}{25} & \coloredaccuracy{39.7}{24} & \coloredaccuracy{42.4}{24} & \coloredaccuracy{63.2}{23} & \coloredaccuracy{47.9}{25} & \coloredaccuracy{68.4}{22} & \coloredaccuracy{16.5}{14} & \coloredaccuracy{46.5}{25} & \coloredaccuracy{28.1}{20} & \coloredaccuracy{69.4}{21} & \coloredaccuracy{89.2}{10} & \coloredaccuracy{65.2}{23} & \coloredaccuracy{54.7}{25} \\

\modelscores{gpt-4o}{43.7}{76.1} & \coloredaccuracy{87.8}{11} & \coloredaccuracy{63.0}{23} & \coloredaccuracy{40.7}{24} & \coloredaccuracy{47.8}{25} & \coloredaccuracy{59.8}{24} & \coloredaccuracy{56.2}{25} & \coloredaccuracy{69.2}{21} & \coloredaccuracy{27.4}{20} & \coloredaccuracy{52.0}{25} & \coloredaccuracy{33.1}{22} & \coloredaccuracy{68.0}{22} & \coloredaccuracy{88.5}{10} & \coloredaccuracy{65.8}{22} & \coloredaccuracy{56.7}{25} \\

\modelscores{gpt-3.5-turbo}{33.9}{62.1} & \coloredaccuracy{74.3}{19} & \coloredaccuracy{49.5}{25} & \coloredaccuracy{31.4}{22} & \coloredaccuracy{34.5}{23} & \coloredaccuracy{66.7}{22} & \coloredaccuracy{43.8}{25} & \coloredaccuracy{50.4}{25} & \coloredaccuracy{13.9}{12} & \coloredaccuracy{41.8}{24} & \coloredaccuracy{26.2}{19} & \coloredaccuracy{55.8}{25} & \coloredaccuracy{70.8}{21} & \coloredaccuracy{59.6}{24} & \coloredaccuracy{48.0}{25} \\

\modelscores{gemma-7b}{19.9}{60.9} & \coloredaccuracy{70.3}{21} & \coloredaccuracy{38.6}{24} & \coloredaccuracy{14.4}{12} & \coloredaccuracy{39.9}{24} & \coloredaccuracy{64.4}{23} & \coloredaccuracy{24.0}{18} & \coloredaccuracy{50.0}{25} & \coloredaccuracy{5.7}{5} & \coloredaccuracy{24.4}{18} & \coloredaccuracy{13.1}{11} & \coloredaccuracy{51.0}{25} & \coloredaccuracy{76.2}{18} & \coloredaccuracy{46.6}{25} & \coloredaccuracy{31.3}{22} \\

\modelscores{code-gemma-7b}{31.2}{66.4} & \coloredaccuracy{87.8}{11} & \coloredaccuracy{48.2}{25} & \coloredaccuracy{30.4}{21} & \coloredaccuracy{30.5}{21} & \coloredaccuracy{51.7}{25} & \coloredaccuracy{47.3}{25} & \coloredaccuracy{55.6}{25} & \coloredaccuracy{10.0}{9} & \coloredaccuracy{40.7}{24} & \coloredaccuracy{21.9}{17} & \coloredaccuracy{58.5}{24} & \coloredaccuracy{85.4}{12} & \coloredaccuracy{57.1}{24} & \coloredaccuracy{44.0}{25} \\

\modelscores{deepseek-coder-1.3b}{14.8}{54.4} & \coloredaccuracy{70.9}{21} & \coloredaccuracy{27.1}{20} & \coloredaccuracy{14.4}{12} & \coloredaccuracy{19.7}{16} & \coloredaccuracy{40.2}{24} & \coloredaccuracy{30.1}{21} & \coloredaccuracy{48.3}{25} & \coloredaccuracy{9.1}{8} & \coloredaccuracy{18.9}{15} & \coloredaccuracy{12.5}{11} & \coloredaccuracy{42.2}{24} & \coloredaccuracy{66.9}{22} & \coloredaccuracy{37.9}{24} & \coloredaccuracy{26.0}{19} \\

\modelscores{deepseek-coder-7b}{23.5}{68.2} & \coloredaccuracy{73.6}{19} & \coloredaccuracy{38.3}{24} & \coloredaccuracy{22.2}{17} & \coloredaccuracy{27.6}{20} & \coloredaccuracy{59.8}{24} & \coloredaccuracy{41.1}{24} & \coloredaccuracy{57.3}{24} & \coloredaccuracy{14.3}{12} & \coloredaccuracy{31.3}{21} & \coloredaccuracy{21.2}{17} & \coloredaccuracy{61.9}{24} & \coloredaccuracy{84.6}{13} & \coloredaccuracy{50.3}{25} & \coloredaccuracy{40.7}{24} \\

\modelscores{granite-code-3b}{11.3}{59.9} & \coloredaccuracy{78.4}{17} & \coloredaccuracy{17.5}{14} & \coloredaccuracy{12.9}{11} & \coloredaccuracy{27.1}{20} & \coloredaccuracy{51.7}{25} & \coloredaccuracy{31.5}{22} & \coloredaccuracy{54.3}{25} & \coloredaccuracy{5.7}{5} & \coloredaccuracy{15.3}{13} & \coloredaccuracy{10.0}{9} & \coloredaccuracy{44.2}{25} & \coloredaccuracy{76.9}{18} & \coloredaccuracy{31.1}{21} & \coloredaccuracy{26.0}{19} \\

\modelscores{granite-code-8b}{16.6}{68.8} & \coloredaccuracy{84.5}{13} & \coloredaccuracy{27.7}{20} & \coloredaccuracy{17.0}{14} & \coloredaccuracy{28.1}{20} & \coloredaccuracy{44.8}{25} & \coloredaccuracy{43.8}{25} & \coloredaccuracy{68.4}{22} & \coloredaccuracy{5.2}{5} & \coloredaccuracy{19.6}{16} & \coloredaccuracy{13.8}{12} & \coloredaccuracy{54.4}{25} & \coloredaccuracy{87.7}{11} & \coloredaccuracy{38.5}{24} & \coloredaccuracy{40.0}{24} \\

\modelscores{code-llama-7b}{18.6}{59.6} & \coloredaccuracy{76.4}{18} & \coloredaccuracy{33.7}{22} & \coloredaccuracy{20.1}{16} & \coloredaccuracy{25.1}{19} & \coloredaccuracy{52.9}{25} & \coloredaccuracy{35.6}{23} & \coloredaccuracy{47.0}{25} & \coloredaccuracy{7.8}{7} & \coloredaccuracy{21.1}{17} & \coloredaccuracy{15.6}{13} & \coloredaccuracy{49.7}{25} & \coloredaccuracy{70.0}{21} & \coloredaccuracy{45.3}{25} & \coloredaccuracy{34.0}{22} \\

\modelscores{code-llama-13b}{21.2}{66.1} & \coloredaccuracy{77.7}{17} & \coloredaccuracy{39.6}{24} & \coloredaccuracy{23.2}{18} & \coloredaccuracy{26.6}{20} & \coloredaccuracy{56.3}{25} & \coloredaccuracy{41.1}{24} & \coloredaccuracy{55.6}{25} & \coloredaccuracy{7.0}{6} & \coloredaccuracy{25.1}{19} & \coloredaccuracy{16.9}{14} & \coloredaccuracy{52.4}{25} & \coloredaccuracy{78.5}{17} & \coloredaccuracy{46.6}{25} & \coloredaccuracy{46.0}{25} \\

\modelscores{llama-2-7b}{3.1}{21.5} & \coloredaccuracy{46.6}{25} & \coloredaccuracy{6.3}{6} & \coloredaccuracy{4.6}{4} & \coloredaccuracy{5.9}{6} & \coloredaccuracy{11.5}{10} & \coloredaccuracy{7.5}{7} & \coloredaccuracy{9.0}{8} & \coloredaccuracy{1.7}{2} & \coloredaccuracy{3.6}{4} & \coloredaccuracy{3.1}{3} & \coloredaccuracy{16.3}{14} & \coloredaccuracy{36.2}{23} & \coloredaccuracy{10.6}{9} & \coloredaccuracy{7.3}{7} \\

\modelscores{llama-2-13b}{6.5}{45.6} & \coloredaccuracy{58.8}{24} & \coloredaccuracy{15.5}{13} & \coloredaccuracy{10.3}{9} & \coloredaccuracy{9.4}{8} & \coloredaccuracy{25.3}{19} & \coloredaccuracy{24.7}{19} & \coloredaccuracy{34.6}{23} & \coloredaccuracy{3.5}{3} & \coloredaccuracy{7.3}{7} & \coloredaccuracy{5.0}{5} & \coloredaccuracy{32.0}{22} & \coloredaccuracy{75.4}{19} & \coloredaccuracy{26.7}{20} & \coloredaccuracy{22.7}{18} \\

\bottomrule
\end{tabular}}
\caption{We cluster feature vectors for all examples in \spider~and~\bird~and compute the mean and variance of each model's correctness per cluster. This yields clusters on which models perform well above their dataset-level execution accuracy (\textcolor{birdcolor}{\texttt{bird}}/\textcolor{spidercolor}{\texttt{spider}}) such as in \cluster{1} and \cluster{12}, while revealing blind spots, or clusters with systematically low performance (\cluster{8} and \cluster{10}). Mapping examples in our \target~(\realistic) to clusters with K-means inference  reveals that the dataset most resembles clusters with weaker performance (Row 3).}
\label{tab:blind-spots}
\end{table*}
We now use the feature space constructed in \S\ref{sec:methodology} to analyze the performance of LLMs when solving NL2SQL problems.
Models must not only be accurate (i.e, predict SQL queries correctly), but also robust (consistently correct in the face of diverse, potentially flawed, or difficult inputs).

While many robustness studies create datasets targeting specific weaknesses~\cite{deng-etal-2021-structure, gan-etal-2021-spider-syn}, they may overlook model robustness gaps on examples with unique combinations of properties.
Building on the ideas in \S\ref{sec:benchmarks}, we cluster examples across datasets and analyze LLM correctness by \textit{cluster} to identify challenging example classes for each model.
Importantly, this analysis supplements the metric of execution accuracy on an entire dataset, which may obfuscate nuanced model behaviors (\S\ref{subsec:motivation}).

\paragraph{Clustering Example Representations.}
We run K-means clustering over the set union of examples in \spider~and~\bird, setting $k=14$ using the elbow method~\cite{thorndike1953belongs}.\footnote{We use the implementation from \url{https://github.com/DistrictDataLabs/yellowbrick}.}
Table~\ref{tab:blind-spots} visualizes the distribution of examples in each cluster along with the raw example counts.

\paragraph{Model Inference.} 
We experiment with 13 instruction-tuned LLMs whose behavior we would like to better understand, making sure to include a mixture of closed and open-source models, code-based and general purpose language models, as well as models of different sizes: 7B code and chat-based \texttt{gemma} models~\cite{team2024gemma}, 7B and 13B code and chat-based \texttt{llama-2} models~\cite{touvron2023llama}, 3B and 8B code \texttt{granite} models~\cite{mishra2024granite}, 1.3B and 7B code \texttt{deepseek} models~\cite{guo2024deepseek}, as well as \texttt{gpt-4o}, \texttt{gpt-4o-mini}, and \texttt{gpt-3.5-turbo}~\cite{hurst2024gpt}.\footnote{
The analyses are model and dataset-agnostic. Others may select any model they want to analyze.
}
We generate predictions for all examples in \spider~and \bird~with all models using Prompt~\ref{prompt:nl2sql} in a zero-shot manner~\cite{gao2023text, yang-etal-2024-synthesizing}, which includes the database schema and three example rows of values.
We report execution accuracy (EX) over the full datasets (Table~\ref{tab:blind-spots}, Column 1) as well as per cluster (Columns 2--15). 

\paragraph{Blind Spots.} All models have varying performance across clusters that, in some cases, significantly deviate from their overall EX (Table~\ref{tab:blind-spots}).
For example, all models perform well on examples in Clusters 1 (\cluster{1}) and 12 (\cluster{12}), averaging well above their overall accuracy on \spider.
Conversely, all models struggled with $\mathbf{C_{8}}$ and \cluster{10}, including the strongest model overall, \texttt{gpt-4o}.

We train a random forest classifier with 100 estimators on all (feature vector, cluster label $\mathbf{C}$) pairs to identify important features for each cluster using mean decrease in impurity.
For example, \cluster{8} contains examples with complex conditional expressions and technical jargon (Table~\ref{tab:cluster-feats}).
We call this a \textit{blind spot} of models, or a class of examples on which a model systematically achieves low performance as compared to its overall accuracy.
In contrast, \cluster{1} contains examples that filter data from a single column of the table and uses aliases for table or column names, requiring minimal additional reasoning to map effectively (Table~\ref{tab:cluster-feats}).

Cluster-based analysis also allows us to identify classes of examples for which cheaper, smaller models perform competitively with larger ones or where open-source models may perform at par with proprietary ones.
For example, \texttt{granite-code-3b} performs on par with \texttt{deepseek-coder-7b} on examples in \cluster{4}, and even exceeds its performance on examples in \cluster{1}.
\texttt{deepseek-coder-1.3b}, our smallest model, outperforms \texttt{granite-code-3b} on \cluster{13}, while performing on par with it on \cluster{14}.
Open-source models like \texttt{code-gemma-7b} perform on par with proprietary models like \texttt{gpt-4o} on examples in \cluster{1}.

Interestingly, we also observe some general-purpose language models exceeding the performance of their code counterparts on certain clusters (e.g. \texttt{gemma-7b} exceeds the performance of \texttt{code-gemma-7b} on \cluster{4} and \cluster{5}, while performing lower on all other clusters). This could indicate that examples in these clusters require linguistic or reasoning skills that general-purpose LLMs may be better at.  
While clusters are dataset-dependent and can be produced with any inference dataset(s), we explicitly propose and will release these example clusters of \bird~and \spider~since they are popular community benchmarks.

\paragraph{Discussion.} This analysis can inform a variety of downstream applications.
Research into the robustness of NL2SQL models may leverage low-performing clusters to build challenge splits of the data, similar in spirit to existing hard splits of natural language inference data~\cite{gururangan-etal-2018-annotation}.
Moreover, blind spots could assist in the \textit{generation} of adversarial test sets of new challenge examples expressing combinations of properties that models systematically struggle with.

Practically, this analysis could help users make informed model selections, since the majority of their own data (\target) may closely resemble only a handful of clusters.
We illustrate this by running K-means inference on \realistic~(\target) to understand which clusters it most closely resembles. Table~\ref{tab:blind-spots} includes the proportion of \realistic~that mapped to each cluster. 
The highest performing cluster, \cluster{11}, includes the smallest amount of \realistic~data, with only 2.6\% of \realistic~examples mapped to it, while the majority of data maps to \cluster{6}, where many models seem to struggle.
Smaller models performing on par with larger, more expensive models in particular clusters that resemble \target~allows practitioners to perform cost-efficient inference.

\section{Question Rewriting}
\label{sec:rewriting}
While \sqlspace~is primarily intended to enable offline analyses or applications (\S\ref{sec:benchmarks} and \S\ref{sec:blind-spots}), we illustrate a query rewriting application that removes features associated with a model's blind spots as a proof-of-concept for an online application.
We pose a scenario where a lightweight correctness estimator alerts a user when their NL2SQL system is likely to produce incorrect SQL for a natural language question.
In this case, a user may be informed that their question can be rewritten \textit{without} a particular feature to increase the chances that the system produces correct SQL.

\paragraph{Correctness Estimation.} For a model $M$ whose performance we seek to improve, we first build a lightweight correctness classifier $C$ to predict whether $M$ will produce correct SQL for an example given its feature vector. 
We obtain binary correctness labels by evaluating predictions from $M$ on all 10,697 examples in \unite~and train a random forest classifier on a subset of data consisting of feature vectors and correctness labels.\footnote{We use 200 estimators, training them on 90\% of \unite~examples and computing feature importance using the remaining 10\%. Table~\ref{tab:correctness-estimator} contains performance metrics.}
Using the remaining set of examples, we compute feature importance with permutation importance~\cite{breiman2001random} on negative examples, giving us an ordering of features in $\mathcal{P}$ most contributing to $C$'s prediction that $M$ will produce incorrect SQL for an example (see Table~\ref{tab:damaging-features} for top predicates for different models).

\begin{table}
\centering
\resizebox{\columnwidth}{!}{

\label{tab:example-features}
\begin{tabular}{ll|cc} 
\toprule
 & \multirow{2}{*}{\textbf{No Rewriting}} & \multicolumn{2}{c}{\textbf{Rewriting}} \\ 
\cline{3-4}
 &  & \texttt{Acc@1} & \texttt{Acc@3} \\ 
\midrule
\texttt{code-gemma-7b} & \multicolumn{1}{c}{$31.2$} & $31.9$ & $\mathbf{37.7}$ \\
\texttt{deepseek-coder-7b} & \multicolumn{1}{c}{$23.5$} & $24.3$ & $\mathbf{30.8}$ \\
\bottomrule
\end{tabular}}
\caption{Rewriting NL questions in \bird~that express features most likely to contribute to an incorrect prediction by a model improves execution accuracy. We report accuracy with the top 1 and 3 negative features (\texttt{acc@1} and \texttt{acc@3}).}
\label{tab:rewrite-results}
\end{table}

\begin{table}
\centering
\small
\resizebox{\columnwidth}{!}{
\begin{tabular}{lcc|cc} 
\toprule
 & \multicolumn{2}{c|}{\texttt{code-gemma-7b-it}} & \multicolumn{2}{c}{\texttt{deepseek-coder-7b-it}} \\
 & \multicolumn{1}{l}{\textbf{Rewrite \checkmarksmall}} & \multicolumn{1}{l|}{\textbf{Rewrite \crossmarksmall}} & \multicolumn{1}{l}{\textbf{Rewrite \checkmarksmall}} & \multicolumn{1}{l}{\textbf{Rewrite \crossmarksmall}} \\
 \midrule
\textbf{Estimator \checkmarksmall} & $22.8$ & $43.7$ & $18.5$ & $51.4$ \\
\textbf{Estimator \crossmarksmall} & $16.0$ & $17.5$ & $15.4$ & $14.7$ \\
\bottomrule
\end{tabular}}
\caption{Percentage breakdown of \bird~examples by correctness estimator prediction accuracy (``estimator right or wrong'') and rewrite success \texttt{Acc@3} (whether rewriting yielded correct SQL) for \texttt{code-gemma-7b-it} and \texttt{deepseek-coder-7b-it}.
}\label{tab:breakdown}
\end{table}

\paragraph{Rewriting with Correctness Estimation.} We select \bird~as the dataset on which we aim to improve model $M$'s performance.
We simulate an online inference scenario with \bird~examples: first, the correctness estimator $C$ predicts whether $M$ will generate correct SQL for example $e$ (calibration plots in Figures~\ref{fig:gemma-calibration} and \ref{fig:deepseek-calibration}).
For predicted failures, we identify top negative features in $e$ using the feature importances we previously computed and prompt 
a \textbf{\texttt{Rewriter}} (here, \texttt{gpt-4o-2024-08-06}) to rewrite the natural language question to address the negative feature it exhibits using Prompt~\ref{prompt:rewrite-feature-based}. 
Here, the \texttt{Rewriter} simulates a human user who may themselves rewrite their question when informed that it contains a feature that may yield incorrect SQL from $M$.\footnote{See Appendix~\ref{appendix:rewriting} for a discussion on feature modulation as well as the cost of rewriting.}
Both the correctness estimator $C$ and the rewriting features are restricted to the 121 features derived from the natural language question and schema, excluding gold SQL-derived features to maintain a realistic inference setting (see Table~\ref{tab:correctness-estimator} for performance metrics of $C$).

We then evaluate the generated SQL conditioned on the rewrite and report EX in two settings: \texttt{Acc@3}, which rewrites the question with the top three negative features and awards $M$ credit if any of the three rewrites produce correct SQL, as well as \texttt{Acc@1}, where we consider only the rewrite conditioned on the top negative feature. 

\paragraph{Results.} We evaluate rewriting on two models: \texttt{code-gemma-7b} and \texttt{deepseek-coder-7b}.
Rewriting questions to remove features that hurt model performance can help improve accuracy (Table~\ref{tab:rewrite-results}), especially when rewrites address multiple negative features (\texttt{Acc@3}).
This indicates that multiple negative features compound to hinder model performance, and rewriting to eliminate just a single negative feature (\texttt{Acc@1}) is often insufficient. 
Rather, comprehensive rewrites addressing multiple negative features that collectively degrade model accuracy yield more substantial improvements (\texttt{Acc@3}).
A realistic setting that leverages this finding may ask a user to eliminate multiple negative features in a single rewrite of their query.

Table~\ref{tab:breakdown} reveals that in most cases, the correctness estimator produced a correct decision (``Estimator \checkmarksmall''), but the generated rewrite did not produce correct SQL (``Rewrite \crossmarksmall''), illustrating that the representation space produced by \sqlspace~is meaningful for modeling correctness, but that a more nuanced procedure to produce better rewrites would likely further improve accuracy, which we leave to future work.

\section{Conclusion}
We present \texttt{SQLSpace}, a framework to construct unified representations of text-to-SQL examples semi-automatically.
\sqlspace~enables detailed and interpretable examination of NL2SQL benchmarks to better understand example properties as well as fine-grained analysis of LLM behavior across benchmarks, and we argue that having such analyses is essential for future benchmark construction and model development in NL2SQL.
We lay the groundwork for several future research directions, such as data augmentation or adversarial example generation informed by model blind spots, the development of an online ``router'' to route difficult examples to larger models, or the design of a more robust, generalizable correctness estimator.

\section*{Limitations}
\paragraph{Human Intervention.} Our goal is to construct a meaningful but compact representation space for text-to-SQL examples with as little manual intervention as possible.
While the human effort necessary to construct these representations is significantly less than other robustness studies that collect manual annotations from database engineers or SQL experts, our semi-automated pipeline still contains two points of manual intervention: determining the five aspects on which to condition description generation, and further predicate filtering.

\paragraph{Correctness Estimator Efficacy.} The performance of the correctness estimators we train during our rewriting experiments varies depending on the dataset, potentially limiting its utility in certain settings.
While the estimators do learn meaningful patterns for the two NL2SQL models we experiment with, it is possible that their ability to estimate the correctness of other NL2SQL models may vary based on the dataset.
Future work could explore more complex models beyond random forests to improve the generalization ability of the correctness estimators, in turn enabling several downstream applications.

\paragraph{Cost.} Several parts of our pipeline utilize closed-source LLMs. While we do this to establish a proof-of-concept, as well as an upper bound of performance, we acknowledge the cost associated with them.
Future work could experiment with replacing the predicator evaluator in \S\ref{sec:methodology} with a lighter-weight, open-source model to enable inference time prediction of features in a new text-to-SQL example.

\bibliography{custom}

\begin{thebibliography}{41}
\expandafter\ifx\csname natexlab\endcsname\relax\def\natexlab#1{#1}\fi

\bibitem[{Angelidis and Lapata(2018)}]{angelidis-lapata-2018-aspect}
Stefanos Angelidis and Mirella Lapata. 2018.
\newblock \href {https://doi.org/10.18653/v1/D18-1403} {Summarizing opinions: Aspect extraction meets sentiment prediction and they are both weakly supervised}.
\newblock In \emph{Proceedings of the 2018 Conference on Empirical Methods in Natural Language Processing}, pages 3675--3686, Brussels, Belgium. Association for Computational Linguistics.

\bibitem[{Artstein and Poesio(2008)}]{artstein2008inter}
Ron Artstein and Massimo Poesio. 2008.
\newblock Inter-coder agreement for computational linguistics.
\newblock \emph{Computational linguistics}, 34(4):555--596.

\bibitem[{Beaver(1997)}]{beaver1997presupposition}
David~Ian Beaver. 1997.
\newblock Presupposition.
\newblock In \emph{Handbook of logic and language}, pages 939--1008. Elsevier.

\bibitem[{Breiman(2001)}]{breiman2001random}
Leo Breiman. 2001.
\newblock Random forests.
\newblock \emph{Machine learning}, 45:5--32.

\bibitem[{Chang et~al.(2023)Chang, Wang, Dong, Pan, Zhu, Li, Lan, Zhang, Jiang, Lilien et~al.}]{chang2023dr}
Shuaichen Chang, Jun Wang, Mingwen Dong, Lin Pan, Henghui Zhu, Alexander~Hanbo Li, Wuwei Lan, Sheng Zhang, Jiarong Jiang, Joseph Lilien, et~al. 2023.
\newblock Dr. spider: A diagnostic evaluation benchmark towards text-to-sql robustness.
\newblock \emph{arXiv preprint arXiv:2301.08881}.

\bibitem[{Deng et~al.(2021)Deng, Awadallah, Meek, Polozov, Sun, and Richardson}]{deng-etal-2021-structure}
Xiang Deng, Ahmed~Hassan Awadallah, Christopher Meek, Oleksandr Polozov, Huan Sun, and Matthew Richardson. 2021.
\newblock \href {https://doi.org/10.18653/v1/2021.naacl-main.105} {Structure-grounded pretraining for text-to-{SQL}}.
\newblock In \emph{Proceedings of the 2021 Conference of the North American Chapter of the Association for Computational Linguistics: Human Language Technologies}, pages 1337--1350, Online. Association for Computational Linguistics.

\bibitem[{Gan et~al.(2021{\natexlab{a}})Gan, Chen, Huang, Purver, Woodward, Xie, and Huang}]{gan-etal-2021-spider-syn}
Yujian Gan, Xinyun Chen, Qiuping Huang, Matthew Purver, John~R. Woodward, Jinxia Xie, and Pengsheng Huang. 2021{\natexlab{a}}.
\newblock \href {https://doi.org/10.18653/v1/2021.acl-long.195} {Towards robustness of text-to-{SQL} models against synonym substitution}.
\newblock In \emph{Proceedings of the 59th Annual Meeting of the Association for Computational Linguistics and the 11th International Joint Conference on Natural Language Processing (Volume 1: Long Papers)}, pages 2505--2515, Online. Association for Computational Linguistics.

\bibitem[{Gan et~al.(2021{\natexlab{b}})Gan, Chen, and Purver}]{gan-etal-2021-spider-dk}
Yujian Gan, Xinyun Chen, and Matthew Purver. 2021{\natexlab{b}}.
\newblock \href {https://doi.org/10.18653/v1/2021.emnlp-main.702} {Exploring underexplored limitations of cross-domain text-to-{SQL} generalization}.
\newblock In \emph{Proceedings of the 2021 Conference on Empirical Methods in Natural Language Processing}, pages 8926--8931, Online and Punta Cana, Dominican Republic. Association for Computational Linguistics.

\bibitem[{Gao et~al.(2023)Gao, Wang, Li, Sun, Qian, Ding, and Zhou}]{gao2023text}
Dawei Gao, Haibin Wang, Yaliang Li, Xiuyu Sun, Yichen Qian, Bolin Ding, and Jingren Zhou. 2023.
\newblock Text-to-sql empowered by large language models: A benchmark evaluation.
\newblock \emph{arXiv preprint arXiv:2308.15363}.

\bibitem[{Grice(1975)}]{grice1975logic}
Herbert~P Grice. 1975.
\newblock Logic and conversation.
\newblock In \emph{Speech acts}, pages 41--58. Brill.

\bibitem[{Guo et~al.(2024)Guo, Zhu, Yang, Xie, Dong, Zhang, Chen, Bi, Wu, Li et~al.}]{guo2024deepseek}
Daya Guo, Qihao Zhu, Dejian Yang, Zhenda Xie, Kai Dong, Wentao Zhang, Guanting Chen, Xiao Bi, Yu~Wu, YK~Li, et~al. 2024.
\newblock Deepseek-coder: When the large language model meets programming--the rise of code intelligence.
\newblock \emph{arXiv preprint arXiv:2401.14196}.

\bibitem[{Gururangan et~al.(2018)Gururangan, Swayamdipta, Levy, Schwartz, Bowman, and Smith}]{gururangan-etal-2018-annotation}
Suchin Gururangan, Swabha Swayamdipta, Omer Levy, Roy Schwartz, Samuel Bowman, and Noah~A. Smith. 2018.
\newblock \href {https://doi.org/10.18653/v1/N18-2017} {Annotation artifacts in natural language inference data}.
\newblock In \emph{Proceedings of the 2018 Conference of the North {A}merican Chapter of the Association for Computational Linguistics: Human Language Technologies, Volume 2 (Short Papers)}, pages 107--112, New Orleans, Louisiana. Association for Computational Linguistics.

\bibitem[{Hazoom et~al.(2021)Hazoom, Malik, and Bogin}]{hazoom-etal-2021-text}
Moshe Hazoom, Vibhor Malik, and Ben Bogin. 2021.
\newblock \href {https://doi.org/10.18653/v1/2021.nlp4prog-1.9} {Text-to-{SQL} in the wild: A naturally-occurring dataset based on stack exchange data}.
\newblock In \emph{Proceedings of the 1st Workshop on Natural Language Processing for Programming (NLP4Prog 2021)}, pages 77--87, Online. Association for Computational Linguistics.

\bibitem[{Hurst et~al.(2024)Hurst, Lerer, Goucher, Perelman, Ramesh, Clark, Ostrow, Welihinda, Hayes, Radford et~al.}]{hurst2024gpt}
Aaron Hurst, Adam Lerer, Adam~P Goucher, Adam Perelman, Aditya Ramesh, Aidan Clark, AJ~Ostrow, Akila Welihinda, Alan Hayes, Alec Radford, et~al. 2024.
\newblock Gpt-4o system card.
\newblock \emph{arXiv preprint arXiv:2410.21276}.

\bibitem[{Kaoshik et~al.(2021)Kaoshik, Patil, Agarawal, Jain, and Singh}]{kaoshik2021aclsql}
Ronak Kaoshik, Rohit Patil, Shaurya Agarawal, Naman Jain, and Mayank Singh. 2021.
\newblock Acl-sql: Generating sql queries from natural language.
\newblock In \emph{Proceedings of the 3rd ACM India Joint International Conference on Data Science \& Management of Data (8th ACM IKDD CODS \& 26th COMAD)}, pages 423--423.

\bibitem[{Lan et~al.(2023)Lan, Wang, Chauhan, Zhu, Li, Guo, Zhang, Hang, Lilien, Hu et~al.}]{lan2023unite}
Wuwei Lan, Zhiguo Wang, Anuj Chauhan, Henghui Zhu, Alexander Li, Jiang Guo, Sheng Zhang, Chung-Wei Hang, Joseph Lilien, Yiqun Hu, et~al. 2023.
\newblock Unite: A unified benchmark for text-to-sql evaluation.
\newblock \emph{arXiv preprint arXiv:2305.16265}.

\bibitem[{Lee et~al.(2021)Lee, Polozov, and Richardson}]{lee-etal-2021-kaggledbqa}
Chia-Hsuan Lee, Oleksandr Polozov, and Matthew Richardson. 2021.
\newblock \href {https://doi.org/10.18653/v1/2021.acl-long.176} {{K}aggle{DBQA}: Realistic evaluation of text-to-{SQL} parsers}.
\newblock In \emph{Proceedings of the 59th Annual Meeting of the Association for Computational Linguistics and the 11th International Joint Conference on Natural Language Processing (Volume 1: Long Papers)}, pages 2261--2273, Online. Association for Computational Linguistics.

\bibitem[{Lei et~al.(2024)Lei, Chen, Ye, Cao, Shin, Su, Suo, Gao, Hu, Yin et~al.}]{lei2024spider}
Fangyu Lei, Jixuan Chen, Yuxiao Ye, Ruisheng Cao, Dongchan Shin, Hongjin Su, Zhaoqing Suo, Hongcheng Gao, Wenjing Hu, Pengcheng Yin, et~al. 2024.
\newblock Spider 2.0: Evaluating language models on real-world enterprise text-to-sql workflows.
\newblock \emph{arXiv preprint arXiv:2411.07763}.

\bibitem[{Li et~al.(2024)Li, Hui, Qu, Yang, Li, Li, Wang, Qin, Geng, Huo et~al.}]{li2024bird}
Jinyang Li, Binyuan Hui, Ge~Qu, Jiaxi Yang, Binhua Li, Bowen Li, Bailin Wang, Bowen Qin, Ruiying Geng, Nan Huo, et~al. 2024.
\newblock Can llm already serve as a database interface? a big bench for large-scale database grounded text-to-sqls.
\newblock \emph{Advances in Neural Information Processing Systems}, 36.

\bibitem[{Ma and Wang(2021)}]{ma2021mt}
Pingchuan Ma and Shuai Wang. 2021.
\newblock Mt-teql: evaluating and augmenting neural nlidb on real-world linguistic and schema variations.
\newblock \emph{Proceedings of the VLDB Endowment}, 15(3):569--582.

\bibitem[{McInnes et~al.(2018)McInnes, Healy, and Melville}]{mcinnes2018umap}
Leland McInnes, John Healy, and James Melville. 2018.
\newblock Umap: Uniform manifold approximation and projection for dimension reduction.
\newblock \emph{arXiv preprint arXiv:1802.03426}.

\bibitem[{Mishra et~al.(2024)Mishra, Stallone, Zhang, Shen, Prasad, Soria, Merler, Selvam, Surendran, Singh et~al.}]{mishra2024granite}
Mayank Mishra, Matt Stallone, Gaoyuan Zhang, Yikang Shen, Aditya Prasad, Adriana~Meza Soria, Michele Merler, Parameswaran Selvam, Saptha Surendran, Shivdeep Singh, et~al. 2024.
\newblock Granite code models: A family of open foundation models for code intelligence.
\newblock \emph{arXiv preprint arXiv:2405.04324}.

\bibitem[{Pi et~al.(2022)Pi, Wang, Gao, Guo, Li, and Lou}]{pi-etal-2022-table}
Xinyu Pi, Bing Wang, Yan Gao, Jiaqi Guo, Zhoujun Li, and Jian-Guang Lou. 2022.
\newblock \href {https://doi.org/10.18653/v1/2022.acl-long.142} {Towards robustness of text-to-{SQL} models against natural and realistic adversarial table perturbation}.
\newblock In \emph{Proceedings of the 60th Annual Meeting of the Association for Computational Linguistics (Volume 1: Long Papers)}, pages 2007--2022, Dublin, Ireland. Association for Computational Linguistics.

\bibitem[{Saparina and Lapata(2024)}]{saparina2024ambrosia}
Irina Saparina and Mirella Lapata. 2024.
\newblock Ambrosia: A benchmark for parsing ambiguous questions into database queries.
\newblock \emph{arXiv preprint arXiv:2406.19073}.

\bibitem[{Searle(1975)}]{searle1975taxonomy}
John~R Searle. 1975.
\newblock A taxonomy of illocutionary acts.

\bibitem[{Sen et~al.(2020)Sen, Lei, Quamar, {\"O}zcan, Efthymiou, Dalmia, Stager, Mittal, Saha, and Sankaranarayanan}]{sen2020fiben}
Jaydeep Sen, Chuan Lei, Abdul Quamar, Fatma {\"O}zcan, Vasilis Efthymiou, Ayushi Dalmia, Greg Stager, Ashish Mittal, Diptikalyan Saha, and Karthik Sankaranarayanan. 2020.
\newblock Athena++ natural language querying for complex nested sql queries.
\newblock \emph{Proceedings of the VLDB Endowment}, 13(12):2747--2759.

\bibitem[{Shi et~al.(2020)Shi, Zhao, Boyd-Graber, Daum{\'e}~III, and Lee}]{shi-etal-2020-squall}
Tianze Shi, Chen Zhao, Jordan Boyd-Graber, Hal Daum{\'e}~III, and Lillian Lee. 2020.
\newblock \href {https://doi.org/10.18653/v1/2020.findings-emnlp.167} {On the potential of lexico-logical alignments for semantic parsing to {SQL} queries}.
\newblock In \emph{Findings of the Association for Computational Linguistics: EMNLP 2020}, pages 1849--1864, Online. Association for Computational Linguistics.

\bibitem[{Srikanth et~al.(2024)Srikanth, Carpuat, and Rudinger}]{srikanth-etal-2024-often}
Neha Srikanth, Marine Carpuat, and Rachel Rudinger. 2024.
\newblock \href {https://doi.org/10.1162/tacl_a_00692} {How often are errors in natural language reasoning due to paraphrastic variability?}
\newblock \emph{Transactions of the Association for Computational Linguistics}, 12:1143--1162.

\bibitem[{Team et~al.(2024)Team, Mesnard, Hardin, Dadashi, Bhupatiraju, Pathak, Sifre, Rivi{\`e}re, Kale, Love et~al.}]{team2024gemma}
Gemma Team, Thomas Mesnard, Cassidy Hardin, Robert Dadashi, Surya Bhupatiraju, Shreya Pathak, Laurent Sifre, Morgane Rivi{\`e}re, Mihir~Sanjay Kale, Juliette Love, et~al. 2024.
\newblock Gemma: Open models based on gemini research and technology.
\newblock \emph{arXiv preprint arXiv:2403.08295}.

\bibitem[{Thorndike(1953)}]{thorndike1953belongs}
Robert~L Thorndike. 1953.
\newblock Who belongs in the family?
\newblock \emph{Psychometrika}, 18(4):267--276.

\bibitem[{Tolo{\c{s}}i and Lengauer(2011)}]{tolocsi2011classification}
Laura Tolo{\c{s}}i and Thomas Lengauer. 2011.
\newblock Classification with correlated features: unreliability of feature ranking and solutions.
\newblock \emph{Bioinformatics}, 27(14):1986--1994.

\bibitem[{Tomova et~al.(2022)Tomova, Hofmann, and M{\"a}der}]{tomova2022seoss}
Mihaela~Todorova Tomova, Martin Hofmann, and Patrick M{\"a}der. 2022.
\newblock Seoss-queries-a software engineering dataset for text-to-sql and question answering tasks.
\newblock \emph{Data in Brief}, 42:108211.

\bibitem[{Touvron et~al.(2023)Touvron, Martin, Stone, Albert, Almahairi, Babaei, Bashlykov, Batra, Bhargava, Bhosale et~al.}]{touvron2023llama}
Hugo Touvron, Louis Martin, Kevin Stone, Peter Albert, Amjad Almahairi, Yasmine Babaei, Nikolay Bashlykov, Soumya Batra, Prajjwal Bhargava, Shruti Bhosale, et~al. 2023.
\newblock Llama 2: Open foundation and fine-tuned chat models.
\newblock \emph{arXiv preprint arXiv:2307.09288}.

\bibitem[{Utama et~al.(2018)Utama, Weir, Basik, Binnig, Cetintemel, H{\"a}ttasch, Ilkhechi, Ramaswamy, and Usta}]{utama2018end}
Prasetya Utama, Nathaniel Weir, Fuat Basik, Carsten Binnig, Ugur Cetintemel, Benjamin H{\"a}ttasch, Amir Ilkhechi, Shekar Ramaswamy, and Arif Usta. 2018.
\newblock An end-to-end neural natural language interface for databases.
\newblock \emph{arXiv preprint arXiv:1804.00401}.

\bibitem[{Wang et~al.(2023{\natexlab{a}})Wang, Gao, Li, and Lou}]{wang-etal-2023-ambigSQL}
Bing Wang, Yan Gao, Zhoujun Li, and Jian-Guang Lou. 2023{\natexlab{a}}.
\newblock \href {https://doi.org/10.18653/v1/2023.findings-acl.352} {Know what {I} don`t know: Handling ambiguous and unknown questions for text-to-{SQL}}.
\newblock In \emph{Findings of the Association for Computational Linguistics: ACL 2023}, pages 5701--5714, Toronto, Canada. Association for Computational Linguistics.

\bibitem[{Wang et~al.(2023{\natexlab{b}})Wang, Shang, and Zhong}]{wang-etal-2023-goal}
Zihan Wang, Jingbo Shang, and Ruiqi Zhong. 2023{\natexlab{b}}.
\newblock \href {https://doi.org/10.18653/v1/2023.emnlp-main.657} {Goal-driven explainable clustering via language descriptions}.
\newblock In \emph{Proceedings of the 2023 Conference on Empirical Methods in Natural Language Processing}, pages 10626--10649, Singapore. Association for Computational Linguistics.

\bibitem[{Yang et~al.(2024)Yang, Hui, Yang, Yang, Lin, and Zhou}]{yang-etal-2024-synthesizing}
Jiaxi Yang, Binyuan Hui, Min Yang, Jian Yang, Junyang Lin, and Chang Zhou. 2024.
\newblock \href {https://doi.org/10.18653/v1/2024.acl-long.425} {Synthesizing text-to-{SQL} data from weak and strong {LLM}s}.
\newblock In \emph{Proceedings of the 62nd Annual Meeting of the Association for Computational Linguistics (Volume 1: Long Papers)}, pages 7864--7875, Bangkok, Thailand. Association for Computational Linguistics.

\bibitem[{Yu et~al.(2019{\natexlab{a}})Yu, Zhang, Er, Li, Xue, Pang, Lin, Tan, Shi, Li, Jiang, Yasunaga, Shim, Chen, Fabbri, Li, Chen, Zhang, Dixit, Zhang, Xiong, Socher, Lasecki, and Radev}]{yu-etal-2019-cosql}
Tao Yu, Rui Zhang, Heyang Er, Suyi Li, Eric Xue, Bo~Pang, Xi~Victoria Lin, Yi~Chern Tan, Tianze Shi, Zihan Li, Youxuan Jiang, Michihiro Yasunaga, Sungrok Shim, Tao Chen, Alexander Fabbri, Zifan Li, Luyao Chen, Yuwen Zhang, Shreya Dixit, Vincent Zhang, Caiming Xiong, Richard Socher, Walter Lasecki, and Dragomir Radev. 2019{\natexlab{a}}.
\newblock \href {https://doi.org/10.18653/v1/D19-1204} {{C}o{SQL}: A conversational text-to-{SQL} challenge towards cross-domain natural language interfaces to databases}.
\newblock In \emph{Proceedings of the 2019 Conference on Empirical Methods in Natural Language Processing and the 9th International Joint Conference on Natural Language Processing (EMNLP-IJCNLP)}, pages 1962--1979, Hong Kong, China. Association for Computational Linguistics.

\bibitem[{Yu et~al.(2018)Yu, Zhang, Yang, Yasunaga, Wang, Li, Ma, Li, Yao, Roman, Zhang, and Radev}]{yu-etal-2018-spider}
Tao Yu, Rui Zhang, Kai Yang, Michihiro Yasunaga, Dongxu Wang, Zifan Li, James Ma, Irene Li, Qingning Yao, Shanelle Roman, Zilin Zhang, and Dragomir Radev. 2018.
\newblock \href {https://doi.org/10.18653/v1/D18-1425} {{S}pider: A large-scale human-labeled dataset for complex and cross-domain semantic parsing and text-to-{SQL} task}.
\newblock In \emph{Proceedings of the 2018 Conference on Empirical Methods in Natural Language Processing}, pages 3911--3921, Brussels, Belgium. Association for Computational Linguistics.

\bibitem[{Yu et~al.(2019{\natexlab{b}})Yu, Zhang, Yasunaga, Tan, Lin, Li, Er, Li, Pang, Chen, Ji, Dixit, Proctor, Shim, Kraft, Zhang, Xiong, Socher, and Radev}]{yu-etal-2019-sparc}
Tao Yu, Rui Zhang, Michihiro Yasunaga, Yi~Chern Tan, Xi~Victoria Lin, Suyi Li, Heyang Er, Irene Li, Bo~Pang, Tao Chen, Emily Ji, Shreya Dixit, David Proctor, Sungrok Shim, Jonathan Kraft, Vincent Zhang, Caiming Xiong, Richard Socher, and Dragomir Radev. 2019{\natexlab{b}}.
\newblock \href {https://doi.org/10.18653/v1/P19-1443} {{SP}ar{C}: Cross-domain semantic parsing in context}.
\newblock In \emph{Proceedings of the 57th Annual Meeting of the Association for Computational Linguistics}, pages 4511--4523, Florence, Italy. Association for Computational Linguistics.

\bibitem[{Yu et~al.(2020)Yu, Chen, Yu, Li, Yang, Jiang, and Jiang}]{yu2020criteria2sql}
Xiaojing Yu, Tianlong Chen, Zhengjie Yu, Huiyu Li, Yang Yang, Xiaoqian Jiang, and Anxiao Jiang. 2020.
\newblock Dataset and enhanced model for eligibility criteria-to-sql semantic parsing.
\newblock In \emph{12th International Conference on Language Resources and Evaluation (LREC)}.

\end{thebibliography}
\clearpage

\appendix
\definecolor{promptblue}{RGB}{181, 179, 242}

\newtcolorbox{code}[1][]{
    fontupper=\footnotesize,
    boxsep=4pt,
    left=0pt,
    right=0pt,
    top=0pt,
    bottom=0pt,
    boxrule=1pt,
    enhanced jigsaw,
    sharp corners,
    #1,
}

\section{Human Intervention in \sqlspace}
\label{appendix:human-intervention}
\sqlspace~discovers features for representation construction in a semi-automatic manner, avoiding a scenario in which database engineers manually come up with perturbations, as in other robustness NL2SQL benchmarks.
\paragraph{Human Intervention in NL2SQL Robustness Studies.}
Many robustness studies rely on significant manual effort.
\citet{pi-etal-2022-table} study whether or not text-to-SQL systems are robust to adversarial perturbations to the tables and schemas involved in text-to-SQL examples. 
Their benchmark took 253 human hours to annotate.
\citet{chang2023dr} introduce \abr{Dr. Spider}, a dataset that introduces 17 human-created perturbations from database experts and crowdworkers.
Here, human intervention involved (1) crowdsourcing annotators from Mechanical Turk to paraphrase questions from \abr{Spider}, (2) task experts (3 SQL experts) that review the paraphrases and categorize them from the text-to-SQL task perspectives, and (3) more task expert hours (3 SQL experts) review the paraphrased questions that are generated from models.
Task experts create ontologies of paraphrases.
Lastly, \citet{gan-etal-2021-spider-syn} study the robustness of text-to-SQL models to synonym substitution. 
They introduce \abr{Spider-Syn}, a human-curated dataset.
Four graduates students in Computer Science  annotate their dataset manually.
The graduate students are trained with a detailed annotation guide, and must to annotate samples in a trial phrase before annotating the whole dataset. 
The annotators manually choose synonyms for substitution in the natural language questions. They have two rounds of annotation, and in total, annotators look at 5,672 questions.

\paragraph{Human Intervention in \sqlspace.}\textbf{In contrast to the related work above, we have no SQL experts or crowdworkers creating any data or ontologies for our framework.} The only manual intervention in \sqlspace~involves 1-2 hours of an author removing semantic duplicates from the generated predicate list in Step 3 of Section~\ref{sec:methodology}. 
This required no task expertise, only proficiency in English.

\section{Ablation Studies}
\label{sec:ablation}
The primary goal of our pipeline is to better characterize and understand features of text-to-SQL datasets in a human-interpretable fashion and how these features help users better understand model performance. 
Consider the four steps in the system:

\begingroup
\addtolength\leftmargini{-0.2in}
\begin{quote}
\vspace{-0.5em}
\small
    [\textbf{Step 1: Describe Examples based on Aspects}] Here, given an \textit{aspect} such as ``natural language syntax'' or ``linguistic pragmatics'', an LLM generates an aspect-conditioned description of a text-to-SQL example.
    
    [\textbf{Step 2: Propose Features (Predicates) from Descriptions}] Here, given a collection of generated descriptions, an LLM can extract short binary natural language predicates that serve as features in \texttt{SQLSpace}.

    [\textbf{Step 3: Remove Duplicate Features}] Using a combination of automatic and manual methods, we clean the list of generated predicates to produce a deduplicated list of features.

    [\textbf{Step 4: Compute Feature Vectors for any Inference Example}] For a given text-to-SQL example $e$, we run through our list of deduplicated features, and use an LLM to predict whether or not the feature is present in $e$ to produce a binary feature vector $v$ that serves as the example's representation for downstream use.
    
\vspace{-0.5em}
\end{quote}
\endgroup

\paragraph{What are the intutions behind each step in the pipeline?} 
The overall goal of SQLSpace is to discover human-interpretable, generalizable features of text-to-SQL examples to use in downstream analysis. 
The overarching intuition is to create a space in which to represent queries, without committing to a prior researcher intuitions.
\textbf{In Step 1}, we ask an LLM to produce a prose description of an example containing commentary related to a particular aspect. 
These descriptions are an intermediate checkpoint before predicate generation. 
One could imagine suggesting features without descriptions just by looking at examples, but this is significantly more challenging for language models since meta-reasoning about all aspects at once is challenging. 
Describing an example before coming up with features is similar to how chain-of-thought serves to improve problem solving in LLMs.

\textbf{Step 2} simply suggests features extracted from descriptions. 
Natural language predicates as features lend  themselves well to the main goal of \sqlspace: human interpretability. Features are short, human-readable, and binary in order to maximize the flexibility and interpretability of our framework. 
This predicate generation step is taken from an explainable clustering pipeline~\cite{wang-etal-2023-goal} and is positioned as a module that performs ``in-context clustering'' based on a random subset of the corpus with explainable cluster properties. 
We use the predicate generation module from their pipeline to produce a list of ``explanations'' of shared properties in our corpus of examples.
This automatic process mirrors a manual process of a database engineer sampling random slices of the example corpus and coming up with properties shared by the majority of examples in the slice.

We employ pruning in \textbf{Step 3} mainly to reduce the dimensionality of the feature vectors and duplicate features, removing paraphrases of features to reduce bloating of the feature set.
A refined version of this step may include the practitioner determining what types of features they are looking to study (assuming a prior). 

\textbf{Step 4} is feature vector construction. Step 3 produces a feature set, and Step 4 builds the feature vector representation for any example by checking whether a feature is ``on'' or ``off'' for an example, as in any feature vector construction in classical machine learning with the exception that \sqlspace~employs an LLM to evaluate whether a feature is ``hot''.

\paragraph{What steps in our pipeline \textit{can} be ablated?} 

We discuss three ablative settings that help illustrate the utility of the decisions made during our pipeline design: (1) predicate proposal from text-to-SQL examples directly \textit{instead} of descriptions which eliminates Step 1 altogether (\ref{ablation:raw-proposal}), (2) aspect-agnostic description generation (\ref{ablation:no-aspects}) which eliminates conditioning on aspects, (3) non-deduplicated predicates (\ref{ablation:no-dedupe}) which eliminates Step 3 altogether.
We compare the predicates produced in these settings to those produced by our final pipeline in \texttt{SQLSpace}.
These ablations revolve around Step 1 (description generation) and Step 3 (predicate deduplication). 
We note that Step 2 (actual generation of predicates) and Step 4 (inference) cannot be materially ablated (turned on or off). 

\subsection{Ablation Analysis 1: Predicate Proposal Directly from Text-to-SQL Examples}
\label{ablation:raw-proposal}

\sqlspace~leverages a predicate proposal module from \citet{wang-etal-2023-goal}, designed to generate a list of explanations, or predicates by prompting an LLM to perform ``in-context clustering''. 
This proposal module ingests a subset of examples from the overall corpus, some natural language goal \textit{g}, and an instruction to produce a set of explanations for the candidate clusters.
Formally, it ingests a sample set $S = \{x_1....x_T\}$, a goal $g$, and an instruction (for example, ``\textit{Generate a list of n explanations for candidate clusters based on the sample set}''). 
See Section 3 of \citet{wang-etal-2023-goal} for more details.

\sqlspace~uses aspect-based descriptions of examples instead of NL2SQL examples themselves as input into the predicate proposal module in order to disentangle \textit{individual} example understanding and reasoning from reasoning required for example \textit{clustering}. 
When generating aspect-based descriptions, an LLM is forced to reason about an example from a particular perspective, and the LLM in the predicate proposal step (Step 2 in \sqlspace) can simply find commonalities among these descriptions.

\paragraph{Goal of Ablation.} To illustrate the utility of this design decision, \textbf{we consider an ablative setting in which we turn off example description generation altogether (turn off Step 1 of \sqlspace)}, and input raw NL2SQL examples directly into the predicate proposal module, forcing the LLM to not only reason about individual examples, but also commonalities across examples.
We analyze the predicates proposed from this setup and compare them to $\mathcal{P}$, the predicate set resulting from first running aspect-based example description generation for each example (Step 1) and inputting these descriptions into the proposal module.

\paragraph{Setup.} We use all NL2SQL examples from \unite~as in \sqlspace~with the same setup as described in Step 2 of Section~\ref{sec:methodology}.
Each sample passed to the proposal LLM look like so:

\begin{code}
\texttt{\textbf{Natural Language Question}:  subject states that he / she has current hepatic disease.
\\
\\
\textbf{SQL}: select id from records where hepatic\_disease = 1}
\end{code}

\noindent We set the goal as following (in contrast to aspect-specific goals in \sqlspace): 

\begin{prompt}[title={Prompt \thetcbcounter: Control Goal}, label=prompt:goal-control]
\texttt{\colorbox{promptblue}{Prompt:} Here are some examples of natural language questions and their corresponding SQL queries. I want to cluster these examples based on similarities.}
\end{prompt}

\paragraph{Results.} This process produced 167 predicates, and after automatically deduplicating with \href{https://github.com/seatgeek/thefuzz}{\texttt{thefuzz}}, we obtain 102 predicates, listed in the first row of Table~\ref{tab:control-predicates}.
Manually inspecting these predicates reveals a few different types of unhelpful properties:

\begin{itemize}
    \item \textbf{Database Content-Based Predicates:} Many predicates revolve around database entities (e.g \texttt{``asks about flights and airlines''}, \texttt{``queries about car details and specifications''}, \texttt{``asks about music albums and songs''}). 
    These are \textit{not} generalizable and are database-specific.
    This is likely due to the proposal LLM is simply finding shallow commonalities in entities in SQL queries.
    Our aspect-based predicates (Table~\ref{tab:full-list}) are designed to focus on meta-properties of examples instead of database content.
    \item \textbf{Generic Properties of SQL Queries:} Proposing predicates from examples themselves without any prior reasoning or example understanding step also yielded predicates that were vague phrases that generally apply to most SQL queries, and are hence not discriminative (e.g. \texttt{``requests specific information based on database criteria''}, \texttt{``seeks details from a database''}, or \texttt{``seeks to retrieve data based on specific criteria''}). 
    These types of predicates do not convey meaningful information about NL2SQL examples, and would be treated as noise if we chose this setting in \sqlspace.
    \item \textbf{Operation Focused:} When predicates were not database-specific or generic/vague, they were mostly focused on the \textit{operation} that the SQL query was trying to perform (e.g. \texttt{``requests for maximum or minimum values''}, \texttt{``requests information about counts or totals''}, \texttt{``asks about comparisons''}).
    While some of our predicates in \sqlspace~are indeed concerned with these types of SQL operations, \sqlspace's final set of predicates contain much richer information about \textit{how} the natural language question and SQL query \textit{achieve} the operation.
\end{itemize}

\textbf{All in all, this ablation experiment highlights the need for a step which explicitly performs the necessary example understanding} for high-quality predicates.
While \sqlspace~is designed for general-purpose use across NL2SQL applications, it is possible that practitioners may indeed want predicates related to topics as a part of their example representation vector. 
In these cases, \sqlspace~offers \textit{flexibility} to describe examples conditioned on whichever aspects practitioners find useful.

\subsection{Ablation Analysis 2: Aspect-Agnostic Description Generation}
\label{ablation:no-aspects}

In Appendix~\ref{ablation:raw-proposal}, we demonstrated that the predicates produced by eliminating an example \textit{understanding} step before predicate proposal yields low-quality, generic, and non-generalizable features.
This helped to illustrate the utility of description generation (Step 1) of the \sqlspace~pipeline described in Section~\ref{sec:methodology}.

Here, we explore another ablative experiment setting of Step 1 that removes \textit{aspect-based} conditioning of description generation for each example (i.e the generation of $d_\text{syn}$, $d_\text{sql-syn}$, $d_\text{sem}$, $d_\text{prag}$, and $d_\text{db}$ from Step 1 in Section~\ref{sec:methodology}), and instead produces general descriptions of each example.

\paragraph{Goal of Ablation.} This ablation helps to demonstrate the utility of conditioning descriptions on a particular aspect.
We analyze the predicates proposed from this setup and again compare them to $\mathcal{P}$ to understand the impact and role of focusing on particular aspects of NL2SQL examples.

\paragraph{Setup.} For each example in \unite, we generate a general description agnostic of any aspect using the prompt below (compare this prompt to Prompts~\ref{prompt:describe-syntax}--\ref{prompt:describe-db}): 

\begin{prompt}[title={Prompt \thetcbcounter: Generating Descriptions: Control}, label=prompt:describe-control]
\texttt{\colorbox{promptblue}{Prompt:} You are an expert SQL programmer and linguist. I will give you a natural language question and the corresponding SQL query. I will also provide the underlying database schema in the form \texttt{\lbrack database\_name\rbrack{} | \lbrack table\rbrack{} : \lbrack column\rbrack{}, \lbrack column\rbrack{} ...| \lbrack table\rbrack{} : ... | ....}. I am a database engineer trying to learn about examples in my dataset. Output a paragraph describing the example.}
\end{prompt}

\noindent We do not instruct the description generator LLM to focus on any aspect of the example, instead letting \textit{it} choose what details and information to include in a summary description of each example.
This process yields high-level descriptions of examples. Compare two descriptions of the same example below: the first description is the aspect-agnostic description, while the second description is $d_\text{syn}$.

\begin{code}
\tiny
\texttt{In this example, we are working with a database named '204\_335' that contains several tables related to tennis tournaments and match outcomes. The tables include 'tournament\_address', 'opponents\_in\_the\_final\_list', 'score\_in\_the\_final\_list', 'score\_in\_the\_final\_list\_number', 'score\_in\_the\_final\_list\_number1', 'score\_in\_the\_final\_list\_number2', and 'w'. Each table has columns that store various details about the matches, such as match IDs, addresses, opponents, scores, and other relevant information. The natural language question asks us to compare the years 1971 and 1972 to determine which year had the most matches won by a player named Osborne. The corresponding SQL query is designed to find this information by selecting the 'date\_number' (which represents the year) from the 'w' table where the 'date\_number' is either 1971 or 1972 and the 'outcome' is 'winner'. The query groups the results by 'date\_number' and orders them by the count of matches won in descending order, limiting the result to the top entry. The SQL query effectively counts the number of matches won by Osborne in each of the specified years and returns the year with the highest count. This approach ensures that we can identify the year in which Osborne had the most victories, providing a clear comparison between the two years in question.}
\end{code}

\begin{code}
\tiny
\texttt{The natural language question "comparing 1971 and 1972, which year has the most won matches by osborne?" presents several interesting syntactic features. Firstly, the question starts with a non-canonical structure, known as "fronting," where the comparative element "comparing 1971 and 1972" is placed at the beginning of the sentence. This fronting not only functions to set the stage for the query but also highlights the focus of the comparison—namely, the two given years. 
Moreover, the question employs a Wh-type clause with "which year," invoking a selection process among a defined set of options (1971 and 1972). This clause acts as the main clause's subject, making the syntactic structure of the sentence an embedded question within a comparison framework. The verb phrase "has the most won matches" maintains subject-verb agreement with the singular subject "year," and showcases a superlative construction ("most") typical in comparative questions.
Additionally, the indirect object "by osborne" follows a prepositional phrase, where "by" denotes the agent performing an action. This prepositional phrase functions syntactically as an adjunct that specifies the performer of the winning matches. The inclusion of the agent within a comparison emphasizes the relationship between the syntactic elements and endows the question with specificity regarding whose matches are being evaluated.
The linguistically intricate layering of a subject-complement (which year), combined with a comparative framework (more), and the specification of the agent (by osborne), all contribute to the syntactic complexity of the natural language question. This structure efficiently narrows down the information scope while maintaining syntactic coherence in a query format.}
\end{code}

\noindent It is worth noting that while the aspect-agnostic description does produce details that are scattered across different aspect-based descriptions, it focuses more on the database content and less on the meta-properties of the examples. 

We take these aspect-agnostic descriptions and feed them into the same predicate proposal module from Step 2 in Section~\ref{sec:methodology}, but again change the natural language goal to be aspect-agnostic, encouraging the model to cluster examples and propose predicates solely based on similarities of generated descriptions: 

\begin{prompt}[title={Prompt \thetcbcounter: Control Goal}, label=prompt:goal-control-descriptions]
\texttt{\colorbox{promptblue}{Prompt:} Here are some descriptions of natural language questions and their corresponding SQL queries. I want to cluster these examples based on similarities.}
\end{prompt}

\paragraph{Results.} This process produced 171 predicates, and after automatically deduplicating with \href{https://github.com/seatgeek/thefuzz}{\texttt{thefuzz}} with a threshold of 85, we obtain 124 predicates, listed in the second row of Table~\ref{tab:control-predicates}.
Inspecting these predicates yields similar results to Ablation~\ref{ablation:raw-proposal}. 
Many predicates are yet again focused on \textit{database content} but to \textbf{an even higher degree}, ignoring \textit{meta-properties} of the example (e.g. \texttt{``concerns retrieving birth dates of tennis players'', ``seeks the number of singles released in a specific year from a music database'' or ``seeks data on poker players' achievements''}). 
These predicates are not generalizable to other databases and therefore do not make for strong features in a unified NL2SQL example representation.
Some predicates are again vague and high-level, applying to all or a majority of SQL examples (\texttt{``utilizes SQL functions and operators'', ``asks for specific information retrieval'', ``focuses on specific data extraction''}).
We do however observe that introducing descriptions, and hence a step that focuses on reasoning about examples, does \textit{start} to introduce meaningful predicates (e.g. \texttt{``demonstrates the use of INTERSECT operator in SQL queries''} or \texttt{``requires joining multiple tables in SQL queries''}), something that was not present at all in Ablation~\ref{ablation:raw-proposal}.

It is exactly this behavior that we sought to encourage with the introduction of aspects in \sqlspace.
However, we still see a focus on predicates related to the SQL query, and very little representation of the linguistic properties of the natural langugage question, an important part of the NL2SQL task.
\textbf{This ablation, taken with the results from the previous ablation setting, help to underscore the importance of explicitly producing aspect-specific conditions, since they allow the model to focus on fine-grained meta-properties of NL2SQL examples that yield high-quality, generalizable, and useful predicates.
}
\subsection{Ablation Analysis 3: No Deduplication}
\label{ablation:no-dedupe}
Lastly, we discuss a setting in which we ablate Step 3 of \sqlspace, predicate deduplication. 
This step was included primarily for cleanliness and avoid noisy or useless features. 
We run the predicate proposal step in \sqlspace (Step 2) with two models: \texttt{gpt-4o-2024-08-06} as well as \texttt{gpt-3.5-turbo-0125}. 
Since both LLMs were provided the same descriptions, many predicates were duplicated. 
Without deduplication, Step 2 produced a total of 1,210 predicates (see breakdown in Table~\ref{tab:proposed-counts}). This would have yielded 1,210-dimensional vectors, a prohibitively high dimension that would decrease interpretability and ease of usage for users of \sqlspace.
Furthermore, duplicate and highly correlated features can affect downstream usage~\cite{tolocsi2011classification}.
This helps illustrate the importance of Step 3, which reduces 1,212 predicates down to a set of just 187, improving interpretability, utility, and removing noise.
Furthermore, Step 3 is essential in reducing the overall latency of \sqlspace, since computing each individual feature for an NL2SQL example requires a call to an LLM. 

\section{Comparing \texttt{SQLSpace} Features to Metadata in \abr{Spider} and \abr{Bird}}
\label{appendix:bird-vs-spider-vs-sqlspace}

\paragraph{Spider.} The authors of \citet{yu-etal-2018-spider} conduct an analysis of the SQL hardness of examples in their dataset to understand dataset composition from a SQL component perspective. 
They divide SQL queries into \textit{easy}, \textit{medium}, \textit{hard}, and \textit{extra hard} solely based on the the number of SQL components, selections and conditions. 
This results in queries with more keywords considered as harder.
These labels are \textbf{static}.

In contrast, \sqlspace~does not prescriptively pre-define what makes an example difficult, but rather finds examples across datasets that share properties, and evaluates models on these groupings.
These groupings can have features that not only span the gold SQL, but also involve properties of the natural language question and its relationship to the underlying schema, allowing us to understand what makes examples difficult \textbf{for a particular model} (clusters where a model performs lower than others). 

\paragraph{Bird.} The authors of \citet{li2024bird} annotate examples as \textbf{simple}, \textbf{moderate}, and \textbf{challenging}, augmenting criteria beyond SQL complexity to include schema linking, reasoning required, and question intent understanding. Here, they ask annotators to provide a judgment on a scale from 1--3 across four dimensions and subsequently rank all examples.
While this setup captures example properties better than in \citet{yu-etal-2018-spider}, it is neither sufficiently fine-grained nor easily interpretable.
A static label of ``simple'' does not offer much insight into example dynamics enough to carefully understand the specific properties that \textit{make} the example simple.

Again, our framework does not pre-define example difficulty, simply identifying features that describe text-to-SQL examples in general, and evaluating models on groups of similar examples.
Our 187-dimensional vector representations of examples allow practitioners to thoroughly understand properties of examples that are simple or challenging for models.

\begin{figure*}[t!]
\centering
\includegraphics[scale=0.5]{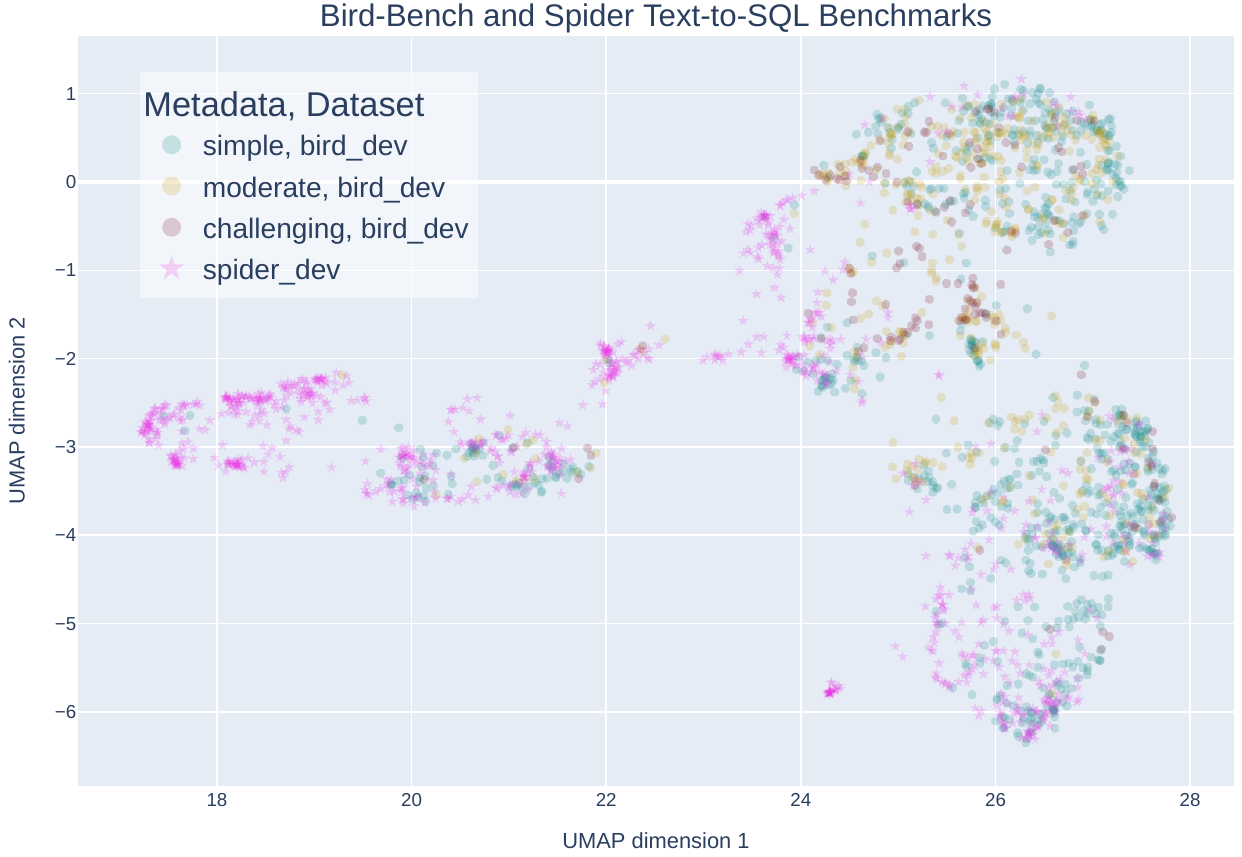}
\caption{\abr{UMAP} projection of feature vectors from \bird~and~\spider. We color each point in \bird~using the hand-annotated metadata released with the dataset. We observe that areas of overlap between \bird~and~\spider~typically occur on examples annotated as ``simple'' in \bird.}
\label{fig:bird-v-spider}
\end{figure*}

\section{Generating Example Descriptions}
\label{appendix:describe-examples}

\begin{prompt}[title={Prompt \thetcbcounter: Generating Descriptions: Syntax}, label=prompt:describe-syntax]
\texttt{\colorbox{promptblue}{Prompt:} You are an expert SQL programmer and linguist. I will give you a natural language question and the corresponding SQL query. I will also provide the underlying database schema in the form \texttt{\lbrack database\_name\rbrack{} | \lbrack table\rbrack{} : \lbrack column\rbrack{}, \lbrack column\rbrack{} ...| \lbrack table\rbrack{} : ... | ....}.  I am a linguist trying to learn about examples in my dataset from a \textbf{linguistic syntax perspective}. This includes anything about word order, grammatical relations, hierarchical sentence structure (constituency), agreement, the nature of cross linguistic variation, and the relationship between form and meaning. Output a highly detailed paragraph describing \textbf{ONLY} fine-grained \textbf{linguistic syntactic} observations about the natural language question.
}
\end{prompt}

\begin{prompt}[title={Prompt \thetcbcounter: Generating Descriptions: SQL Syntax}, label=prompt:describe-sql-syntax]
\texttt{\colorbox{promptblue}{Prompt:} You are an expert SQL programmer and linguist. I will give you a natural language question and the corresponding SQL query. I will also provide the underlying database schema in the form \texttt{\lbrack database\_name\rbrack{} | \lbrack table\rbrack{} : \lbrack column\rbrack{}, \lbrack column\rbrack{} ...| \lbrack table\rbrack{} : ... | ....}. I am a database engineer trying to learn about examples in my dataset from a \textbf{SQL syntax perspective}. Specifically, I would like to learn about the \textbf{structure} of the SQL query, the \textbf{complexity} of the query, the \textbf{relationship between the query and the provided underlying database schema}, and the nature of cross-database variation. Output a highly detailed paragraph describing \textbf{ONLY} fine-grained observations about the SQL query.
}
\end{prompt}

\begin{prompt}[title={Prompt \thetcbcounter: Generating Descriptions: Example Semantics}, label=prompt:describe-example-semantics]
\texttt{\colorbox{promptblue}{Prompt:} You are an expert SQL programmer and linguist. I will give you a natural language question and the corresponding SQL query. I will also provide the underlying database schema in the form \texttt{\lbrack database\_name\rbrack{} | \lbrack table\rbrack{} : \lbrack column\rbrack{}, \lbrack column\rbrack{} ...| \lbrack table\rbrack{} : ... | ....}. I am a database engineer trying to learn about examples in my dataset of SQL queries. Specifically, I would like to learn about the \textbf{relationship} between the provided natural language question and the SQL query. For example, how does the natural language question relate to the SQL query? Do they exhibit parallel characteristics, or is there some reasoning required to map between the two? What kind of reasoning is required? What are the similarities and differences between the two? Output a highly detailed paragraph describing \textbf{ONLY} fine-grained \textbf{comparison-based} observations about the natural language question versus the SQL query.}
\end{prompt}

\begin{prompt}[title={Prompt \thetcbcounter: Generating Descriptions: Pragmatics}, label=prompt:describe-pragmatics]
\texttt{\colorbox{promptblue}{Prompt:} You are an expert SQL programmer and linguist. I will give you a natural language question and the corresponding SQL query. I will also provide the underlying database schema in the form \texttt{\lbrack database\_name\rbrack{} | \lbrack table\rbrack{} : \lbrack column\rbrack{}, \lbrack column\rbrack{} ...| \lbrack table\rbrack{} : ... | ....}. I am a linguist trying to learn about examples in my dataset from a linguistic pragmatics perspective. Specifically, I would like to learn about the pragmatics of the natural language question. For example, what speech acts are used in the question? Include commentary on Gricean theory, implicature, relevance, and any other information about how word choice and context contribute to the meaning. Does the question exhibit vagueness, underspecification, or ambiguity that make it difficult to understand the author's intent? Output a highly detailed paragraph describing \textbf{ONLY} fine-grained linguistic pragmatic observations about the natural language question.}
\end{prompt}

\begin{prompt}[title={Prompt \thetcbcounter: Generating Descriptions: Database Reasoning}, label=prompt:describe-db]
\texttt{\colorbox{promptblue}{Prompt:} You are an expert SQL programmer and linguist. I will give you a natural language question and the corresponding SQL query. I will also provide the underlying database schema in the form \texttt{\lbrack database\_name\rbrack{} | \lbrack table\rbrack{} : \lbrack column\rbrack{}, \lbrack column\rbrack{} ...| \lbrack table\rbrack{} : ... | ....}. The provided natural language question is attempting to access information from the provided database schema. I am a database engineer and I want to learn about the \textbf{relationship} between the natural language question and the provided database schema. To what degree is the question grounded in the schema? Does the question use exact column names from the schema? Do the concepts and need expressed in the question have clear counterparts in the database schema? If not, what types of reasoning are required to map between the two? Explain what kind of reasoning is required. For example, is linguistic reasoning required (e.g. analogical reasoning, syntactic reasoning or paraphrastic reasoning)? Is commonsense reasoning required? Is logical reasoning required (e.g. deductive reasoning or causal reasoning). How does the \textbf{structure} of the question (syntactic or semantic) relate to the structure of the database schema? Output a highly detailed paragraph describing \textbf{ONLY} these sorts of fine-grained observations about the relationship between the \textbf{natural language question} and the provided \textbf{database schema}.
}
\end{prompt}

\section{Predicate Discovery}
\label{appendix:predicate-discovery}

\begin{prompt}[title={Prompt \thetcbcounter: Predicate Discovery: Syntax}, label=prompt:goal-syn]
\texttt{\colorbox{promptblue}{Prompt:} Here are some detailed descriptions of natural language questions and their corresponding SQL queries. I want to cluster these descriptions based on observations about \textbf{linguistic syntax}.}
\end{prompt}

\begin{prompt}[title={Prompt \thetcbcounter: Predicate Discovery: SQL Syntax}, label=prompt:goal-sql-syn]
\texttt{\colorbox{promptblue}{Prompt:} Here are some detailed descriptions of natural language questions and their corresponding SQL queries. I want to cluster these descriptions based on observations about the \textbf{syntax of the SQL query}.}
\end{prompt}

\begin{prompt}[title={Prompt \thetcbcounter: Predicate Discovery: Example Semantics}, label=prompt:goal-sem]
\texttt{\colorbox{promptblue}{Prompt:} 
Here are some detailed descriptions of natural language questions and their corresponding SQL queries. I want to cluster these descriptions based on \textbf{comparisons between the provided natural language question and the SQL query.}
}
\end{prompt}

\begin{prompt}[title={Prompt \thetcbcounter: Predicate Discovery: Pragmatics}, label=prompt:goal-prag]
\texttt{\colorbox{promptblue}{Prompt:} 
Here are some detailed descriptions of natural language questions and their corresponding SQL queries. I want to cluster these descriptions based on observations about linguistic pragmatics.
}
\end{prompt}

\begin{prompt}[title={Prompt \thetcbcounter: Predicate Discovery: Database Reasoning}, label=prompt:goal-db]
\texttt{\colorbox{promptblue}{Prompt:} 
Here are some detailed descriptions of natural language questions written to query an underyling database schema. I want to cluster these descriptions based on the relationship and reasoning between the provided natural language question and the underyling database schema.
}
\end{prompt}

\begin{table}
\centering
\label{tab:example-features}
\resizebox{\columnwidth}{!}{
\begin{tabular}{lcc|c} 
\toprule
 & \multicolumn{3}{c}{\textbf{\# of Proposed Predicates }} \\ 
\cline{2-4}
\textbf{Aspect} & \texttt{gpt-4o} & \texttt{gpt-3.5-turbo} & \textbf{Total} \\ 
\midrule
\textbf{Syntax} & 100 & 130 & 230 \\
\textbf{SQL Syntax} & 102 & 127 & 229 \\
\textbf{Example Semantics} & 160 & 171 & 331 \\
\textbf{Pragmatics} & 98 & 117 & 215 \\
\textbf{Database Reasoning} & 171 & 134 & 305 \\
\bottomrule
\end{tabular}}
\caption{We propose candidate predicates for each aspect using both \texttt{gpt-4o} and \texttt{gpt-3.5-turbo} in the predicate proposal step. Each model proposes $n$ predicates in an iterative fashion. The number of predicates proposed by both models are shown above, stratified by aspect.}
\label{tab:proposed-counts}
\end{table}

\section{Predicate Evaluation}

\begin{prompt}[title={Prompt \thetcbcounter: Predicate Evaluation: Syntax}, label=prompt:pred-eval-syntax]
\texttt{\colorbox{promptblue}{Prompt:} 
You will be given some text. Determine whether the TEXT satisfies a PROPERTY. Respond with Yes or No. When uncertain, output No.
\\
\\
Now complete the following example:
\\
\\
PROPERTY: \textbf{{property}}
\\
TEXT: \textbf{{question}}
\\
\\
Does the \textbf{{text}} exhibit the \textbf{PROPERTY}?:}
\end{prompt}

\begin{prompt}[title={Prompt \thetcbcounter: Predicate Evaluation: SQL Syntax}, label=prompt:pred-eval-syntax]
\texttt{\colorbox{promptblue}{Prompt:} 
You will be given a SQL query and a PROPERTY. Determine whether the SQL query satisfies the PROPERTY. Respond with Yes or No. When uncertain, output No.
\\
\\
Now complete the following example:
\\
\\
PROPERTY: \textbf{{property}}
\\
SQL Query: \textbf{{query}}
\\
\\
Does the \textbf{{query}} exhibit the \textbf{PROPERTY}?:}
\end{prompt}

\begin{prompt}[title={Prompt \thetcbcounter: Predicate Evaluation: Semantics}, label=prompt:pred-eval-sem]
\texttt{\colorbox{promptblue}{Prompt:} 
You will be given a natural language question and its SQL translation. You will also be given a PROPERTY. Determine whether the natural language question and its SQL translation satisfy the PROPERTY. Respond with Yes or No. When uncertain, output No.
\\
\\
Now complete the following example:
\\
\\
PROPERTY: \textbf{{property}}
\\
NATURAL LANGUAGE QUESTION: \textbf{{question}}
\\
SQL Query: \textbf{{query}}
\\
\\
Does the question and its SQL translation exhibit the \textbf{PROPERTY}?:}
\end{prompt}

\begin{prompt}[title={Prompt \thetcbcounter: Predicate Evaluation: Pragmatics}, label=prompt:pred-eval-prag]
\texttt{\colorbox{promptblue}{Prompt:} 
You will be given a natural language question and its SQL translation. You will also be given a PROPERTY. Determine whether the natural language question and its SQL translation satisfy the PROPERTY. Respond with Yes or No. When uncertain, output No.
\\
\\
Now complete the following example:
\\
\\
PROPERTY: \textbf{{property}}
\\
NATURAL LANGUAGE QUESTION: \textbf{{question}}
\\
SQL Query: \textbf{{query}}
\\
\\
Does the question and its SQL translation exhibit the \textbf{PROPERTY}?:}
\end{prompt}

\begin{prompt}[title={Prompt \thetcbcounter: Predicate Evaluation: Database Reasoning}, label=prompt:pred-eval-db]
\texttt{\colorbox{promptblue}{Prompt:} 
You will be given a natural language question that is trying to query a database as well as the database schema. You will also be given a PROPERTY. Determine whether the natural language question satisfies the PROPERTY. Respond with Yes or No. When uncertain, output No.
\\
\\
Now complete the following example:
\\
\\
PROPERTY: \textbf{{property}}
\\
DATABASE SCHEMA: \textbf{{question}}
\\
QUESTION: \textbf{{question}}
\\
\\
Does the natural language question exhibit the \textbf{PROPERTY} with respect to the database schema?}
\end{prompt}

\begin{table}
\centering
\resizebox{\columnwidth}{!}{
\begin{tabular}{lcc} 
\toprule
 & \multicolumn{1}{l}{} & \multicolumn{1}{l}{} \\
\textbf{Aspect} & \begin{tabular}[c]{@{}c@{}}\textbf{Mean}\\\textbf{Accuracy}\end{tabular} & \begin{tabular}[c]{@{}c@{}}\textbf{Annotator~}\\\textbf{Agreement ($\kappa$)}\end{tabular} \\ 
\midrule
Syntax & $67.0$ & $41.6$ \\
SQL Syntax & $86.0$ & $81.6$ \\
Example Semantics & $77.0$ & $61.4$ \\
Pragmatics & $63.0$ & $49.0$ \\
Database Reasoning & $72.0$ & $54.5$ \\
\bottomrule
\end{tabular}}
\caption{The average accuracy of \texttt{gpt-4o} on predication evaluation for a sample of 250 example-predicate pairs, stratified by aspect. We also compute the Cohen's Kappa between the two annotator judgments.}
\label{tab:predicate-eval}
\end{table}

\section{NL2SQL Inference}

\begin{prompt}[title={Prompt \thetcbcounter: NL2SQL Inference}, label=prompt:nl2sql]
\texttt{\colorbox{promptblue}{Prompt:} Write an SQLite query to answer the following question given the database schema and example rows. Please wrap your code answer using \textasciigrave\textasciigrave\textasciigrave: 
\\
Schema: \textbf{\{schema\} }
\\
Question: \textbf{\{question\}}
\\
Write a SQLite query wrapped in \textasciigrave\textasciigrave\textasciigrave to answer the question and output nothing else:}
\end{prompt}

\begin{table*}[t!]
\centering
\resizebox{2.08\columnwidth}{!}{
\begin{tabular}{l|l} 
\toprule
\multicolumn{1}{c|}{\textbf{\texttt{code-gemma-7b}}} & \multicolumn{1}{c}{\texttt{\textbf{deepseek-coder-7b}}} \\ 
\hline
requires domain-specific knowledge & involves temporal constraints in the natural language question \\
involves semantic mapping from NL to DB & employs punctuation for clarity \\
adheres to Gricean conversational maxims & contains vague or ambiguous language \\
requires commonsense reasoning to understand the question & question is~clearly grounded in the provided database schema \\
has a slight vagueness in meaning, & uses quantifiers for specificity \\
\bottomrule
\end{tabular}}
\caption{Top five most discriminative features associated with a model $M$ producing incorrect SQL. We use permutation importance to derive an ordered list of all predicates in $\mathcal{P}$. During rewriting, we remove (or add in) the most important feature that the example expresses.}
\label{tab:damaging-features}
\end{table*}
\begin{table}
\centering
\small
\resizebox{\columnwidth}{!}{
\begin{tabular}{@{}l@{\hspace{2pt}}ccccc|cccc@{}} 
\toprule
 & & \multicolumn{4}{c|}{\textbf{\unite}} & \multicolumn{4}{c}{\textbf{\bird}} \\
 & & P & R & F1 & Acc & P & R & F1 & Acc \\ 
\midrule
\multirow{2}{*}{\rotatebox[origin=c]{90}{\scriptsize\textbf{offline}}} 
& \texttt{\scriptsize gemma} & $79.0$ & $82.1$ & $80.5$ & $81.1$ & $52.4$ & $73.7$ & $61.2$ & $70.9$ \\
& \texttt{\scriptsize deepseek} & $79.7$ & $74.4$ & $77.0$ & $80.4$ & $40.1$ & $63.2$ & $49.0$ & $70.0$ \\
\midrule
\multirow{2}{*}{\rotatebox[origin=c]{90}{\scriptsize\textbf{online}}} 
& \texttt{\scriptsize gemma} & $71.0$ & $76.3$ & $73.5$ & $73.6$ & $46.5$ & $50.3$ & $48.3$ & $66.4$ \\
& \texttt{\scriptsize deepseek} & $70.4$ & $66.5$ & $68.4$ & $73.1$ & $37.1$ & $40.2$ & $38.6$ & $69.9$ \\
\bottomrule
\end{tabular}}
\caption{\small Precision, recall, F1, and accuracy of the random forest correctness estimator for both \texttt{code-gemma-7b-it} and \texttt{deepseek-coder-7b-it}. We include results from both the offline version of the correctness estimator that has access to the full set of $\mathcal{P}$ features, as well as the online version that has access to only the 121 features related to the natural language question and schema. \unite~Test is the 10\% holdout set used for feature importances.}
\label{tab:correctness-estimator}
\end{table}

\section{Question Rewriting}
\label{appendix:rewriting}

\paragraph{Feature Modulation.} Features can be helpful (``question is grounded in the database schema'') or harmful (``has vagueness in meaning'') to the NL2SQL task. Negative examples may express bad features, in which case we want to \textit{remove} them during the rewrite, or be absent of positive features, in which case we want to \textit{add} them during the rewrite. We modulate this mode during prompting based on manual assignments to features. In our proposed setting, a human rewriting would naturally modulate this mode themselves.

\paragraph{Cost of Rewriting.} Section~\ref{sec:rewriting} poses an ``online'' scenario in which for an inference example consisting of only a natural language question and the underlying database schema, (1) a feature vector is computed for the example using only the set of features extractable from the question and schema alone (121 features), (2) a lightweight correctness estimator ingests this feature vector and outputs a binary prediction on whether or not the text-to-SQL model is likely to produce correct SQL for the question, and (3) if not, alerts the user to rewrite their query based on the blind-spot features present in the question.
In our simulation of this scenario, we use an LLM to rewrite the question.

Constructing the feature vector from the natural language question and database schema is the step requiring the most compute resources, since an LLM evaluates whether or not each of the 121 features is present in the example or not. 
Our experiments leverage \texttt{gpt-4o} as the feature evaluator. 
While we do not know the number of parameters in \texttt{gpt-4o}, we send concurrent API requests for each of the 121 features to the model, reducing the overhead of feature vector computation down to an average of 10-15 seconds per example.
Future work may explore the possibility of using a smaller LLM (e.g 1B---4B parameters) and sending batched inference calls for feature evaluation, which would likely dramatically decrease the overhead in an online setting.

We consider two settings in which we feel use of rewriting can be justified.
First, consider a scenario in which we are trying to boost the performance of \textbf{a particularly \textit{weak} model} on text-to-SQL (e.g. a model \textit{not} trained on code).
As demonstrated by our experiments in Section~\ref{appendix:rewriting}, rewriting natural language questions to remove negative features can help boost performance. 
Rewriting questions to optimize the performance of a weak text-to-SQL model can improve accuracy without the need to retrain or finetune.
Second, consider a scenario in which we are trying to boost the performance of \textbf{a particular \textit{large model}} on text-to-SQL (e.g. a model with $>$ 100B parameters).
Here, the cost of each call can greatly outweigh the cost of feature vector construction.
Targeted natural language question rewrites can help optimize the performance of the model on text-to-SQL without excessive calls to the large model.

\begin{figure*}
     \centering
     \begin{subfigure}[b]{0.4\textwidth}
         \centering
         \includegraphics[width=\textwidth]{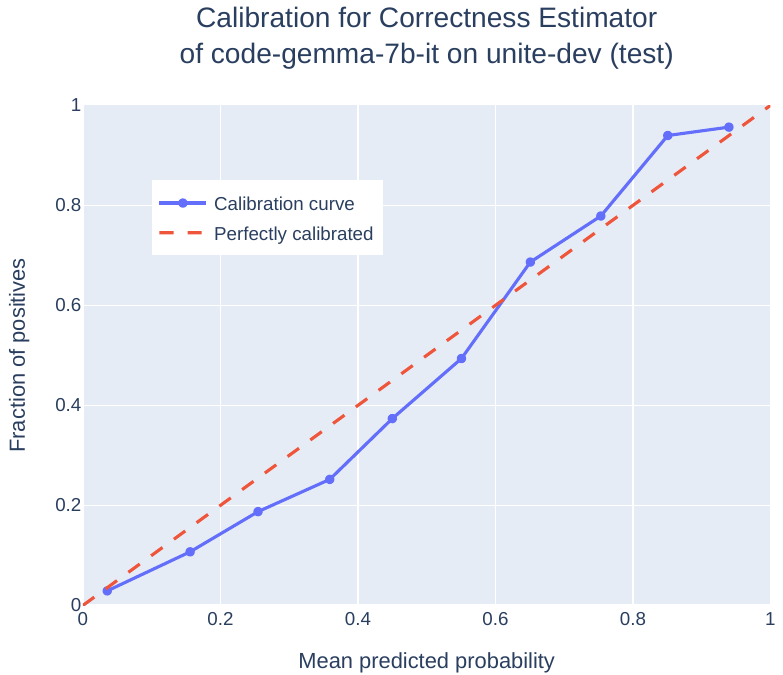}
         \vspace{-1em}
         \caption{Confidence Calibration Plot for \texttt{code-gemma-7b} on the held-out 10\% test split of \unite.}
         \label{fig:gemma-unite-calibration}
     \end{subfigure}
     \hspace{1.5em}
     \begin{subfigure}[b]{0.4\textwidth}
         \centering
\includegraphics[width=\textwidth]{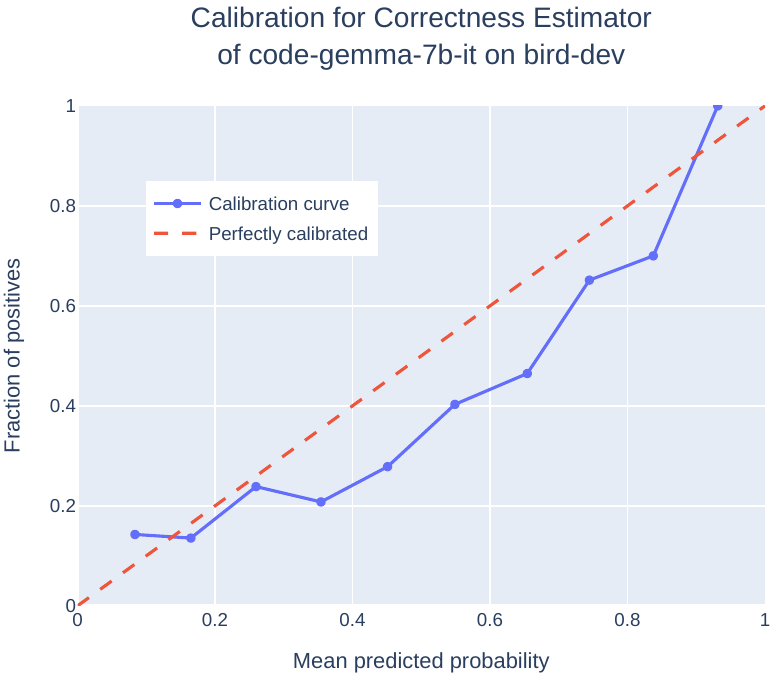}
         \vspace{-1em}
         \caption{Confidence Calibration Plot for \texttt{code-gemma-7b} on \bird.}
         \label{fig:gemma-bird-calibration}
     \end{subfigure}
     \caption{Visualizing calibration of the random forest classifiers we train to predict whether or not \texttt{code-gemma-7b} will produce correct SQL given a compact representation of the example.}
     \label{fig:gemma-calibration}
     \vspace{-1.5em}
     \hfill
\end{figure*}

\begin{figure*}
     \centering
     \begin{subfigure}[b]{0.4\textwidth}
         \centering
         \includegraphics[width=\textwidth]{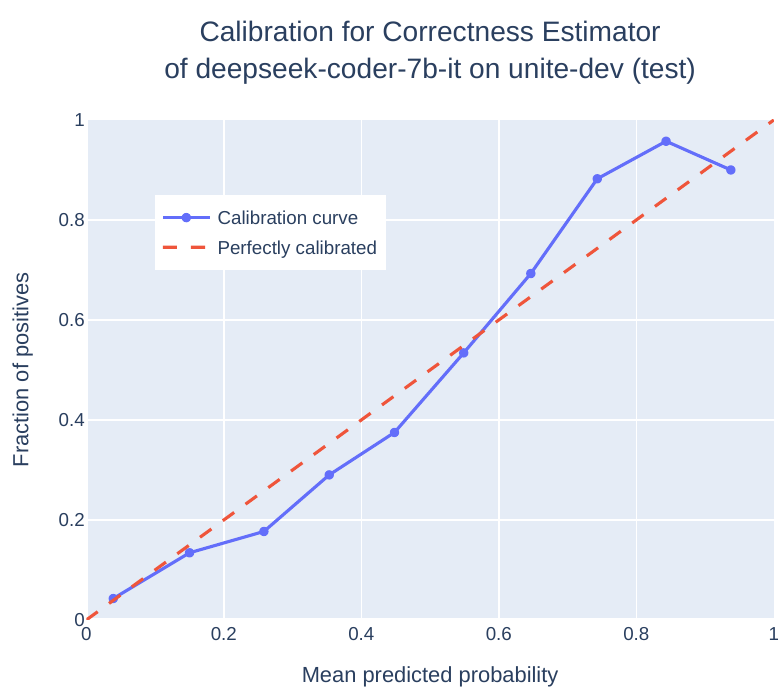}
         \vspace{-1em}
         \caption{Confidence Calibration Plot for \texttt{deepseek-coder-7b} on the held-out 10\% test split of \unite.}
         \label{fig:deepseek-unite-calibration}
     \end{subfigure}
     \hspace{1.5em}
     \begin{subfigure}[b]{0.4\textwidth}
         \centering
\includegraphics[width=\textwidth]{figures/code-gemma-7b-it-calibration-online-bird-dev.pdf}
         \vspace{-1em}
         \caption{Confidence Calibration Plot for \texttt{deepseek-coder-7b} on \bird.}
         \label{fig:deepseek-bird-calibration}
     \end{subfigure}
     \caption{Visualizing calibration of the random forest classifiers we train to predict whether or not \texttt{deepseek-coder-7b} will produce correct SQL given a compact representation of the example.}
     \label{fig:deepseek-calibration}
     \vspace{-1.5em}
     \hfill
\end{figure*}

\begin{prompt}[title={Prompt \thetcbcounter: Rewriting with Natural Language-based Features}, label=prompt:rewrite-feature-based]
\texttt{\colorbox{promptblue}{Prompt:} Given the definition of \textbf{\{feature\}}, I want to rewrite the following natural language question to {mode} \textbf{\{feature\}}. I want the meaning and intent of the question to be preserved. The question I want to rewrite is \textbf{\{question\}}
\\
\\
It is trying to query a database with this schema: \textbf{\{schema\}}. Only output your rewritten question and wrap it in ```. Your question must be as detailed as possible. DO NOT drop information from the original question. If the question cannot be rewritten with the property, output ``INVALID''.
\\
\\
Rewritten \textbf{semantically equivalent} natural language question that expresses \textbf{\{feature\}}:}
\end{prompt}

\section{Computational Resources and Hyperparameters}
We run inference for all open-source models on two NVIDIA A6000 GPUs.
We set a temperature of 0.7 for description generation, and a temperature of 0.1 during NL2SQL generation.

\onecolumn  
\begin{center}
\tiny
\begin{longtable}{p{0.2\textwidth}p{0.7\textwidth}}  
\caption{Top 10 most important features computed with permutation importance for all clusters, consisting of examples from both \spider~and~\bird.}\label{tab:cluster-feats} \\

\toprule
 & \multicolumn{1}{c}{\begin{tabular}[c]{c}\textbf{Permutation Importance-Based}\\\textbf{Top Features}\end{tabular}} \\ 
\midrule
\endfirsthead

\multicolumn{2}{c}{Table \thetable{} (continued)} \\
\toprule
 & \multicolumn{1}{c}{\begin{tabular}[c]{c}\textbf{Permutation Importance-Based}\\\textbf{Top Features}\end{tabular}} \\ 
\midrule
\endhead

\bottomrule
\multicolumn{2}{r}{Continued on next page...} \\
\endfoot

\bottomrule
\endlastfoot

\cluster{1} & \begin{tabular}[t]{p{0.8\textwidth}}involves logical inference beyond direct keyword matching (0.06)\\filters data from a single column of the table (0.06)\\contains a single condition in the WHERE clause (0.05)\\directly aligns with a specific column in the schema, requiring minimal additional reasoning to map effectively (0.04)\\requires additional contextual understanding (0.03)\\involves reasoning about data representation (0.03)\\contains a WHERE clause (0.03)\\uses aliases for table or column names (0.03)\\requires commonsense reasoning to understand the question (0.03)\textcolor[rgb]{0.576,0.631,0.631}{}\end{tabular} \\
\midrule
\cluster{2} & \begin{tabular}[t]{p{0.8\textwidth}}contains a single condition in the WHERE clause (0.07)\\filters data based on a single criterion (0.06)\\retrieves data from multiple tables (0.06)\\uses a JOIN operation (0.05)\\uses aliases for table or column names (0.03)\\utilizes prepositional phrases for modification (0.03)\\uses prepositional phrases to modify noun phrases (0.03)\\utilizes nominal phrases with modifying prepositional phrases (0.03)\\uses modifiers to specify conditions (0.03)\\employs a WHERE clause with multiple conditions (0.02)\end{tabular} \\
\midrule
\cluster{3} & \begin{tabular}[t]{p{0.8\textwidth}}uses subordinate clauses for postmodification (0.06)\\contains subordinate clauses modifying main clauses (0.05)\\contains subordinate clauses (0.05)\\contains nested clauses (0.04)\\contains relative clauses (0.04)\\contains subordinate clauses modifying noun phrases (0.04)\\employs relative clauses to modify noun phrases (0.03)\\uses complex noun phrases (0.02)\\involves the use of logical operators to define selection criteria (0.02)\\contains a WHERE clause (0.01)\end{tabular} \\
\midrule
\cluster{4} & \begin{tabular}[t]{p{0.8\textwidth}}uses a LIMIT clause (0.11)\\includes an ORDER BY clause (0.07)\\limits the selection based on a specific column condition (0.07)\\contains a WHERE clause (0.06)\\involves comparison operations (0.03)\\employs a GROUP BY clause (0.03)\\retrieves data from multiple tables (0.02)\\uses a JOIN operation (0.02)\\uses aliases for table or column names (0.02)\\filters data based on a single criterion (0.02)\end{tabular} \\
\midrule
\cluster{5} & \begin{tabular}[t]{p{0.8\textwidth}}includes a HAVING clause (0.2)\\employs a GROUP BY clause (0.07)\\contains a WHERE clause (0.05)\\utilizes comparative numerals and inequality symbols (0.04)\\exhibits conditional logic (0.03)\\utilizes range specification in nominal phrases (0.02)\\directly maps threshold values to their respective columns in the SQL query (0.02)\\includes an ORDER BY clause (0.02)\\has a complex conditional expression (0.02)\\employs comparison operations for numerical values (0.02)\end{tabular} \\
\midrule
\cluster{6} & \begin{tabular}[t]{p{0.8\textwidth}}filters data based on a single criterion (0.05)\\retrieves data from multiple tables (0.05)\\filters data from a single column of the table (0.05)\\uses a JOIN operation (0.04)\\contains a single condition in the WHERE clause (0.04)\\uses aliases for table or column names (0.03)\\involves missing or incomplete information in either natural language question or SQL query (0.02)\\requires commonsense reasoning to understand the question (0.02)\\clearly grounded in the provided database schema, with no additional reasoning required (0.02)\\contains vague or ambiguous language (0.01)\end{tabular} \\
\midrule
\cluster{7} & \begin{tabular}[t]{p{0.8\textwidth}}limits the selection based on a specific column condition (0.08)\\contains a WHERE clause (0.07)\\has a more detailed filtering component in the SQL query (0.05)\\requires domain-specific knowledge (0.04)\\follows a basic SELECT-FROM-WHERE structure (0.04)\\involves reasoning about data representation (0.03)\\involves logical inference beyond direct keyword matching (0.02)\\involves interpretation and understanding of context (0.02)\\requires understanding of logical operations and condition grouping (0.02)\\contains a single condition in the WHERE clause (0.02)\end{tabular} \\
\midrule
\cluster{8} & \begin{tabular}[t]{p{0.8\textwidth}}contains a CASE statement (0.06)\\has a complex conditional expression (0.04)\\exhibits minimal vagueness or ambiguity (0.03)\\adheres to Gricean conversational maxims (0.03)\\involves stricter conditions than the natural language question (0.02)\\uses aggregate functions in the SELECT clause (0.02)\\contains technical jargon (0.02)\\requires understanding of logical operations and condition grouping (0.02)\\primarily based on direct mappings with some synonymy and domain-specific knowledge required (0.02)\\closely grounded in the database schema, although it does not use exact column names (0.02)\end{tabular} \\
\midrule
\cluster{9} & \begin{tabular}[t]{p{0.8\textwidth}}employs a WHERE clause with multiple conditions (0.1)\\involves the use of logical AND operators (0.09)\\contains a single condition in the WHERE clause (0.05)\\uses comparison operators to connect conditions in the WHERE clause (0.03)\\filters data based on a single criterion (0.03)\\has missing constraints in the sql query (0.02)\\contains vague or ambiguous language (0.02)\\contains relative clauses (0.02)\\underspecifies information (0.02)\\exhibits conditional logic (0.02)\end{tabular} \\
\midrule
\cluster{10} & \begin{tabular}[t]{p{0.8\textwidth}}includes nested sub-conditions (0.05)\\involves the use of logical operators to define selection criteria (0.04)\\utilizes range specification in nominal phrases (0.04)\\employs logical quantifiers to set multiple criteria (0.04)\\involves the use of logical AND operators (0.04)\\includes numerical comparison with inequality symbols (0.04)\\utilizes range specifications (0.04)\\integrates quantified expressions within a conditional framework (0.03)\\employs a WHERE clause with multiple conditions (0.03)\\employs comparison operations for numerical values (0.03)\end{tabular} \\
\midrule
\cluster{11} & \begin{tabular}[t]{p{0.8\textwidth}}employs logical connectors and operators (0.08)\\uses logical operators to connect clauses (0.07)\\uses coordinating conjunctions (0.05)\\employs coordinated noun phrases to express relationships (0.04)\\uses conjunctions to coordinate noun phrases (0.03)\\involves the use of logical operators to define selection criteria (0.03)\\uses coordinating conjunctions for enumeration (0.03)\\employs coordinated noun phrases (0.03)\\employs logical quantifiers to set multiple criteria (0.02)\\employs alternative conditions (0.02)\end{tabular} \\
\midrule
\cluster{12} & \begin{tabular}[t]{p{0.8\textwidth}}requires recognition of implicit relationships between natural language concepts (0.08)\\limits the selection based on a specific column condition (0.06)\\involves logical inference beyond direct keyword matching (0.06)\\involves interpretation and understanding of context (0.05)\\breaks down the problem into specific conditions that need to be met (0.05)\\contains a WHERE clause (0.05)\\demonstrates specificity in language use (0.05)\\involves reasoning about data representation (0.04)\\directly aligns with a specific column in the schema, requiring minimal additional reasoning to map effectively (0.03)\\involves understanding of specific technical terms and concepts (0.03)\end{tabular} \\
\midrule
\cluster{13} & \begin{tabular}[t]{p{0.8\textwidth}}contains a single condition in the WHERE clause (0.06)\\uses prepositional phrases to modify noun phrases (0.04)\\utilizes nominal phrases with modifying prepositional phrases (0.04)\\filters data based on a single criterion (0.04)\\utilizes prepositional phrases for modification (0.04)\\uses a JOIN operation (0.02)\\utilizes prepositional phrases to modify nouns (0.02)\\retrieves data from multiple tables (0.02)\\employs a WHERE clause with multiple conditions (0.02)\\uses aliases for table or column names (0.02)\end{tabular} \\
\midrule
\cluster{14} & \begin{tabular}[t]{p{0.8\textwidth}}contains subqueries (0.15)\\has a nested subquery (0.15)\\has a nested SELECT statement (0.13)\\involves nested logic (0.06)\\uses a nested WHERE clause (0.06)\\uses a single select statement (0.04)\\uses aliases for table or column names (0.02)\\retrieves data from multiple tables (0.01)\\uses a JOIN operation (0.01)\\uses multiple nested parentheses (0.01)\end{tabular} \\

\end{longtable}
\end{center}
\twocolumn  

\onecolumn  
\begin{center}
\tiny
\begin{longtable}{p{0.1\textwidth}p{0.7\textwidth}}  
\caption{Final set of 187 predicates used as features in \texttt{SQLSpace} stratified by aspect.}\label{tab:full-list} \\

\toprule
 & \multicolumn{1}{c}{\textbf{Predicates}} \\ 
\midrule
\endfirsthead

\multicolumn{2}{c}{Table \thetable{} (continued)} \\
\toprule
 & \multicolumn{1}{c}{\begin{tabular}[c]{c}\textbf{Permutation Importance-Based}\\\textbf{Top Features}\end{tabular}} \\ 
\midrule
\endhead

\bottomrule
\multicolumn{2}{r}{Continued on next page...} \\
\endfoot

\bottomrule
\endlastfoot

\textbf{Syntax} & uses complex sentence structures, involves conditional statements, uses prepositional phrases to modify noun phrases, contains subordinate clauses modifying main clauses, has a compound sentence structure, utilizes prepositional phrases for modification, has a conditional clause, employs passive voice constructions, contains elided predicates, contains subordinate clauses, contains subordinate clauses modifying noun phrases, employs hierarchically structured sentence elements, contains multiple coordinate phrases joined by conjunctions, contains relative clauses, uses prepositional phrases to indicate hierarchy, contains a passive voice construction with a predicate adjective, contains subordinate clauses with multiple conditions, uses subordinate clauses for postmodification, uses coordinating conjunctions for enumeration, includes nested sub-conditions, contains nested clauses, employs ellipsis, uses mathematical symbols and units within the syntactic structure, integrates quantified expressions within a conditional framework, uses complex prepositional phrases, employs non-restrictive modifiers, incorporates post-modifier adjective phrases, contains nested prepositional phrases, employs technical lexicon, employs parenthetical expressions, employs punctuation for clarity, demonstrates parallel syntactic constructions, uses shorthand notation, exhibits truncated, telegraphic style, utilizes comparative numerals and inequality symbols, employs mathematical symbols and units, uses punctuation to parse complex syntactic units, uses non-standard shorthand notation for numerical range, includes mixed use of symbols and words, uses shorthand notation to indicate range, employs non-standard linguistic constructions, employs nominal phrases for specification, employs coordinated noun phrases, uses complex noun phrases, employs relative clauses to modify noun phrases, uses conjunctions to coordinate noun phrases, utilizes prepositional phrases to modify nouns, employs coordinated noun phrases to express relationships, includes modifiers for noun phrases, employs nominal phrases for subject and object, utilizes nominal phrases with modifying prepositional phrases, employs nominal phrases as subjects, utilizes range specification in nominal phrases, employs nominal phrases with elided predicates, uses typographic symbols to modify nouns, involves the use of disjunctions, employs logical connectors and operators, uses logical operators to connect clauses, uses comparison operations for selection criteria, uses coordinating conjunctions, uses conjunctions to connect clauses, has a clear logical flow, involves comparison operations, involves the use of logical operators to define selection criteria, employs logical quantifiers to set multiple criteria, utilizes range specifications, employs logical quantifiers like 'and' and 'or', employs comparison operations for numerical values, employs Boolean expressions conjoined with logical operators, uses conjunctions to express disjunction, employs inclusive disjunction with 'or' coordinating conjunction, uses modifiers to specify conditions, includes numerical comparison with inequality symbols \\
\midrule
\textbf{SQL Syntax} & has a nested subquery, has a nested SELECT statement, uses a nested WHERE clause, contains subqueries, employs a LEFT JOIN, employs a LEFT OUTER JOIN, uses a JOIN operation, contains multiple joins, uses a self join, contains a cross join operation, utilizes a common table expression (CTE), uses logical OR operators, uses multiple nested parentheses, employs a GROUP BY clause, uses a union operator, includes aggregate functions, contains a WHERE clause, employs a WHERE clause with multiple conditions, includes a HAVING clause, follows a basic SELECT-FROM-WHERE structure, includes a window function, uses aggregate functions in the SELECT clause, limits the selection based on a specific column condition, contains a CASE statement, filters data based on a single criterion, contains a single condition in the WHERE clause, uses a LIMIT clause, involves the use of logical AND operators, employs a recursive common table expression, contains a pivot or unpivot operation, uses a correlated subquery, uses comparison operators to connect conditions in the WHERE clause, uses a single select statement, includes an ORDER BY clause, contains a UNION ALL operator, uses aliases for table or column names, retrieves data from multiple tables, filters data from a single column of the table, uses a distinct keyword, has a complex conditional expression \\
\midrule
\textbf{Example Semantics} & requires understanding of logical operations and condition grouping, involves understanding of specific technical terms and concepts, involves semantic mapping, requires domain-specific knowledge, involves parallel characteristics between natural language and sql query, involves logical inference beyond direct keyword matching, requires mapping of natural language concepts to database schema, involves stricter conditions than the natural language question, has a direct relationship with no reasoning required, has a more detailed filtering component in the SQL query, directly maps threshold values to their respective columns in the SQL query, involves interpretation and understanding of context, requires recognition of implicit relationships between natural language concepts, involves reasoning about unique identifiers, involves reasoning about data representation, requires identifying table and column names, has missing constraints in the sql query, requires technical familiarity with the database schema, requires additional contextual understanding, involves complex logical reasoning, involves missing or incomplete information in either natural language question or SQL query, requires mapping of high-level concepts to database schema, breaks down the problem into specific conditions that need to be met, involves nested logic, requires recognizing present state vs. past action, involves temporal constraints in the natural language question
\\
\midrule
\textbf{Pragmatics} & uses quantifiers for specificity, relies on context for relevance, has a clear and specific request, employs direct speech acts, has a slight vagueness in meaning, adheres to Gricean conversational maxims, relies on conversational implicature, demonstrates assertive speech act, invokes epistemic modality, contains technical jargon, has a cooperative intention, uses declarative structure for assertive statement, demonstrates specificity in language use, employs declarative statements as questions, underspecifies information, exhibits conditional logic, employs reported speech, seeks specific and measurable information, contains vague or ambiguous language, performs an indirect speech act, uses disjunction to allow for multiple criteria, employs alternative conditions, shows brevity at the cost of quantity maxim, assumes prior knowledge, relies on presuppositions, presumes mutual understanding of context, exhibits minimal vagueness or ambiguity, uses rhetorical questions, demonstrates careful word choice, demonstrates a delicate balance of precision and interpretive flexibility, underspecified in terms of database schema, implies a causal relationship
  \\
\midrule
\textbf{Database Reasoning} & requires commonsense reasoning to understand the question, requires semantic mapping between natural language and schema, requires both syntactic and semantic reasoning to bridge the question's semantics to the schema's structure, involves both synonymy and analogical mapping to align natural language terms with schema column names, requires commonsense reasoning to understand the implicit subject of the table entries, primarily based on direct mappings with some synonymy and domain-specific knowledge required, requires basic deductive logic to map between the natural language question and the schema, clearly grounded in the provided database schema, with no additional reasoning required, mirrors the structure of the database schema, with little room for ambiguity or interpretation, directly aligns with a specific column in the schema, requiring minimal additional reasoning to map effectively, uses exact column names from the schema to query for specific values, involves paraphrastic reasoning to map natural language expressions to schema-specific terminology, is already expressed in a highly structured and database-oriented way, requiring no additional reasoning to map to the schema, uses syntactic variation to describe a condition, closely grounded in the database schema, although it does not use exact column names, semantically aligns with the database schema, but requires some degree of linguistic and commonsense reasoning to accurately map the concepts
\\

\end{longtable}
\end{center}
\twocolumn  

\onecolumn  
\begin{center}
\tiny
\begin{longtable}{p{0.1\textwidth}p{0.7\textwidth}}  
\caption{Example descriptions used to mine predicates for \texttt{SQLSpace}.}\label{tab:description-examples} \\

\toprule
 & \multicolumn{1}{c}{\textbf{Predicates}} \\ 
\midrule
\endfirsthead

\multicolumn{2}{c}{Table \thetable{} (continued)} \\
\toprule
 & \multicolumn{1}{c}{\begin{tabular}[c]{c}\textbf{Permutation Importance-Based}\\\textbf{Top Features}\end{tabular}} \\ 
\midrule
\endhead

\bottomrule
\multicolumn{2}{r}{Continued on next page...} \\
\endfoot

\bottomrule
\endlastfoot

\textbf{Syntax} & \texttt{The natural language question  Find the number of orchestras whose record format is 'CD' or 'DVD'' displays a variety of syntactic features relevant to linguistic analysis. Firstly, the sentence employs a relative clause ('whose record format is 'CD' or 'DVD'') that modifies the noun 'orchestras.' This relative clause specifies a condition related to 'orchestras,' showcasing typical English syntactic behavior where the relative pronoun 'whose' introduces additional information and establishes a relationship between 'orchestras' and 'record format.'  The main clause 'Find the number of orchestras' contains an imperative verb 'Find,' which signals a direct command. This imperative construction lacks an explicit subject, a common feature in English commands where the subject 'you' is understood implicitly.  The noun phrase 'the number of orchestras' consists of a definite article 'the,' a noun 'number,' and a prepositional phrase 'of orchestras,' which together form a well-defined noun phrase indicating a specific quantity. The prepositional phrase functions as a postmodifier to specify what kind of 'number' is being requested. Additionally, the relative clause 'whose record format is 'CD' or 'DVD'' features a coordination structure 'CD or DVD,' highlighting a binary choice within the linguistic structure. The use of 'is' within the relative clause indicates singular agreement with 'record format,' demonstrating subject-predicate agreement.  The relative clause further relies on the copular verb 'is' followed by the disjunctive conjunction 'or,' linking two alternatives in a linear syntactic arrangement, which presents a straightforward binary condition. Overall, the sentence structure exemplifies several critical syntactic phenomena: the use of imperative mood, relative clause formation, noun phrase construction, subject-predicate agreement, and coordination within a relative clause, all contributing to an intricate yet coherent inquiry.} \\
\midrule
\textbf{SQL Syntax} & \texttt{The given SQL query is relatively straightforward and has a simple structure, which includes a SELECT statement aimed at retrieving a single column value from a specific table. The query's complexity is minimal because it involves only one table and a straightforward WHERE clause without any JOIN operations, nested queries, or aggregate functions. In terms of the relationship between the query and the provided underlying database schema from '204\_206', the SQL operation specifically targets the 'w' table. The `SELECT` clause chooses the `time\_greatest\_utc` column, indicating that the objective is to retrieve the time in UTC format for a specific date. The `WHERE` clause filters the rows of the table where the `date` column matches 'october 3, 2415'.  This implies a one-to-one mapping between the question's focus on a specific date and the database's date column, showing a direct translation of the natural language input to SQL command. Moreover, no type casting or formatting functions are employed, hinting that the provided date format in the query is directly comparable to the stored data. In terms of cross-database variation, this example uses a fairly common SQL syntax that should be universally understood across different SQL-compliant databases like MySQL, PostgreSQL, SQLite, and others without requiring modifications. This universality is due to its use of basic SQL keywords (SELECT, FROM, WHERE) and standard date format handling.  Overall, the SQL statement is simple and efficient for this specific query because it directly addresses the requirement using the basic querying capabilities of SQL and is highly readable to those familiar with standard SQL operations.} \\
\midrule
\textbf{Example Semantics} & \texttt{In the provided example, the natural language question directly translates to the corresponding SQL query with a high degree of parallelism. The question specifies 'how many automobiles,' which maps to the SQL function \`count(*)\`, indicating a need to count rows. The term 'were produced' implies a filter condition, which in SQL is represented by the `WHERE` clause. The specific year '1980' in the question is directly used in the SQL query's condition `YEAR = 1980`. Thus, there is a clear one-to-one correspondence between the elements in the natural language and SQL query. The natural language inherently asks for a count based on production year, and the SQL query represents this request accurately without requiring any complex reasoning or transformation. Both the question and the query focus on the same entity (automobiles) and reference the same attribute (year), showing a straightforward and direct mapping with parallel structure and semantics. The differences lie mainly in syntax and format rather than in conceptual understanding or logical structure.}
\\
\midrule
\textbf{Pragmatics} & \texttt{The natural language question 'uncontrolled hypertension' displays several interesting facets from a pragmatics perspective. Firstly, it is not presented in the form of a typical question, exhibiting an elliptical style which omits the interrogative structure commonly expected in querying databases. This ellipsis can signal a form of implicit performative speech act, where the speaker's intent is to request information or specify a condition without using a direct question. According to Grice's Maxims, the utterance violates the Maxim of Manner which prefers clarity and the avoidance of ambiguity since the phrase is underspecified and lacks a clear syntactic structure. The Maxim of Quantity is also in question here, as the statement assumes a shared understanding of what is meant by 'uncontrolled' despite not explicitly mentioning it in the context, thereby creating some potential for ambiguity. This could implicature that the speaker had previously established a context where the control level of hypertension was discussed, although it is unexpressed here. Furthermore, the Maxim of Relevance suggests that within the context of the database schema provided, 'uncontrolled hypertension' is pertinent to 'hypertension' being coded as '1', implicating a shorthand reference within a well-understood domain-specific language. The choice of the term 'uncontrolled' implies a subjective, clinical threshold which is not necessarily informationally equivalent to the binary schema indicated in the database but relies on an implied understanding. Therefore, the phrase exhibits a form of semantic underspecification, leaving it open to interpretation whether 'uncontrolled' is directly equatable to 'hypertension = 1'. This concise expression's interpretive relies heavily on the presumed shared knowledge and context between the issuer of the query and the interpreter.}
  \\
\midrule
\textbf{Database Reasoning} & \texttt{The natural language question, 'What are the country codes of countries where people use languages other than English?' is well-aligned with the provided database schema for the most part. The question and schema share direct counterparts: 'country codes' corresponds to the 'CountryCode' column in the 'countrylanguage' table, and 'languages other than English' is explicitly related to the 'Language' column in the same table. The phrasing 'languages other than English' requires a form of paraphrastic reasoning to translate into the SQL syntax `LANGUAGE != 'English'`, which involves a negation operation in the WHERE clause. Additionally, the word 'distinct' in SQL query (`SELECT DISTINCT CountryCode`) is derived from understanding that multiple countries might have the same code listed under various languages, necessitating uniqueness. Therefore, both syntactic/semantic matching and an element of logical reasoning are required here to achieve the correct SQL query. The structure of the question semantically corresponds to the schema as it clearly asks about the diverse attribute values within specific columns, effectively tapping into the database's relational structure.}\\

\end{longtable}
\end{center}
\twocolumn  

\onecolumn  
\begin{center}
\tiny
\begin{longtable}{p{0.1\textwidth}p{0.7\textwidth}}  
\caption{Predicates from ablation studies.}\label{tab:control-predicates} \\

\toprule
 & \multicolumn{1}{c}{\textbf{Predicates}} \\ 
\midrule
\endfirsthead

\multicolumn{2}{c}{Table \thetable{} (continued)} \\
\toprule
 & \multicolumn{1}{c}{\begin{tabular}[c]{c}\textbf{Permutation Importance-Based}\\\textbf{Top Features}\end{tabular}} \\ 
\midrule
\endhead

\bottomrule
\multicolumn{2}{r}{Continued on next page...} \\
\endfoot

\bottomrule
\endlastfoot

\textbf{NL2SQL Examples (Ablation~\ref{ablation:raw-proposal})} & requests information about entities and their attributes, inquires about top results or rankings, requests data related to issues and resolutions, asks for information related to specific entities or categories, asks for comparisons between different data points, inquires about specific data based on criteria, inquires about specific details or attributes in a dataset, requests for maximum or minimum values, inquires about the maximum or minimum values within a dataset, asks for counts or frequencies, seeks information about the highest or lowest values, inquires about database queries related to entities, requests counts or sums based on specific conditions, requests information about counts or totals, seeks specific details based on conditions, asks about the popularity or rankings of certain items, asks about flights and airlines, seeks to compare different data points, asks for comparisons between different data entries, inquires about database statistics, queries about car details and specifications, inquires about the number of occurrences or counts, requests specific information based on numerical values, requests information about rankings or top results, requests specific data filtering criteria, seeks specific information with numerical results, asks about music albums and songs, asks about comparisons, seeks details on issues and resolutions, requests specific information based on database criteria, requests specific details about data based on certain criteria, requests comparisons between different entities, seeks aggregate data based on certain criteria, seeks details from a database, requests statistical analysis, requests comparison between different data points, inquires about the highest or lowest values, seeks to identify the top or bottom entries based on a specific attribute, seeks to identify extremes or limits, requires filtering based on multiple conditions, inquires about numerical values, inquires about numerical comparisons, asks for statistical calculations, requests information based on conditional criteria, "requests specific details about a single entity, involves comparison between different entities, seeks to find the maximum/minimum value, asks about singer details, involves querying for specific information, inquires about rankings or superlatives, inquires about comparative data, requests information about rankings or top values, inquires about relationships between entities, inquires about TV series and ratings, requests for the count of occurrences based on certain conditions, requests information about rankings or extremes, asks for counts or statistics from a database, inquires about specific data filtering criteria, asks for details on winners and rankings, asks for comparisons between different entities, seeks specific information about car details, inquires about flight details, seeks specific information based on comparisons, seeks details on issue tracking and resolution, requests counts or totals of data, requests specific information about data based on conditions, seeks details about particular entities, involves querying for statistical information, inquires about specific data based on conditions, requests counts or sums based on certain criteria, asks for comparisons or rankings, inquires about winners and rankings, seeks details about maximum or minimum values, requests information based on aggregate functions, asks about the count or sum of certain database entries, seeks details on data related to winners and rankings, inquires about specific data points, "seeks aggregated statistical information, seeks information about relationships between different entities, requests information on document templates and ids, seeks data related to multiple categories, asks for counts or totals, inquires about rankings or ordering, seeks aggregated results from multiple entities, inquires about data related to specific categories, requests information on issues and resolutions, requests statistical summaries, requests information on car-related data, requests information on flights and airlines, seeks to filter and display distinct data entries, seeks aggregate information, asks for details on specific attributes or characteristics, asks for statistical analysis, seeks details about document templates, seeks comparisons between different data entries, asks about flight details and statistics, asks for information related to TV series, requests information about the maximum or minimum value in a dataset, seeks to retrieve data based on specific criteria, asks about the number of occurrences based on certain conditions, seeks information on contestants and votes, inquires about rankings or extremes
 \\
\midrule
\textbf{Aspect-Agnostic Descriptions (Ablation~\ref{ablation:no-aspects})} & involves correcting errors in SQL queries, retrieves specific information about poker players from a database, concerns retrieving birth dates of tennis players, addresses a specific issue ID in a database, involves querying for specific information from databases, involves querying for data based on specific conditions, requests data related to poker player performance, inquires about states with both guardians and veterinarians, seeks the number of singles released in a specific year from a music database, asks for the birth date of a player with a specific name, pertains to identifying authors affiliated with a specific organization and working in a particular field, seeks a list of affiliates with the number of authors working in a specific field, pertains to querying player details from a tennis database, involves listing affiliates and the number of authors working in a specific field, seeks information on authors in a particular field and their affiliates, involves counting entries based on a condition in a database, asks for specific information about a player, asks for a count based on a specific condition, involves querying for specific player details in a tennis database, pertains to counting singles released in a specific year, involves querying multiple tables with joins, inquires about the number of authors in a specific field per affiliate, targets a specific issue and retrieves related information, focuses on counting authors in a specific field per affiliate, concentrates on listing affiliates and authors in a field, seeks information about authors affiliated with a particular organization and field, targets specific data retrieval based on provided criteria, seeks data on poker players' achievements, queries for fix version of a particular software issue, asks for details about final tables made and best finishes of poker players, involves retrieving author names based on affiliations and fields of study, demonstrates the importance of accurate SQL query formulation, utilizes SQL functions for data retrieval, asks for the fix version of an issue in a software engineering database, queries for specific details related to software development, requests information about a specific issue in software development, queries for the birth date of a specific tennis player in a sports database, inquires about authors affiliated with 'PRESTO Japan Science and Technology Corporation' working in a specific field, inquires about the presence of specific groups in certain states, utilizes multiple table joins for data retrieval, focuses on joining multiple tables in a database, retrieves data based on certain conditions, targets a specific issue's fix version in a software engineering database, utilizes filtering based on conditions, asks for details about authors and their affiliations, requests data on singles released in a specific year, focuses on database schema and tables, involves querying for birth date based on player name, inquires about authors affiliated with a specific organization and field of study, requires joining multiple tables in SQL queries, seeks fix version information for a software issue in an SEOSS database, queries for states with specific types of residents, requests a list of affiliates in a particular field, asks for details about poker players' performance and achievements, asks for the birth date of a player named Justine, involves counting entries based on specific criteria, emphasizes the importance of accurate SQL query translations, focuses on identifying common attributes between tables, inquires about poker player performance metrics, joins multiple tables to retrieve data, seeks information about authors affiliated with specific organizations and fields of study, inquires about the number of singles released in a particular year, addresses fixing versions of specific software issues, utilizes SQL functions and operators, requests specific data from a database related to academic research, pertains to final tables made and best finishes for poker players, concerns finding states with specific groups of residents, seeks information on singles released in a specific year, focuses on joining multiple tables in sql queries, aims to retrieve authors affiliated with a specific organization and field from an academic database, inquires about the fix version of a specific issue, requests the fix version of a specific issue in a software development database, filters data based on specific conditions, requires correcting errors in SQL queries, seeks data related to tennis players and their rankings, focuses on querying for specific information from databases, queries for final tables made and best finishes of poker players in a poker database, demonstrates the importance of accurate SQL query representation, addresses errors in SQL queries for accurate data retrieval, requests counts based on specific criteria, focuses on retrieving information about authors and their affiliations, requests data on final tables made and best finishes for poker players, involves finding common data points between two tables in a dog kennel database, requests the count of entries based on a specific criterion, asks for specific information about tennis players, requests information about poker player performance, inquires about fix versions for software development issues, involves working with database tables and columns, queries for specific information from a database, requests the count of authors working in a specific field per affiliate, focuses on querying for player data in a database, asks for specific information retrieval, involves joining multiple tables in SQL queries, seeks names of authors affiliated with a specific organization in a particular field, requests information on fix versions for software development issues, seeks information about music releases in a specific year, emphasizes the importance of accurate conditions in the WHERE clause, retrieves fix version of a specific issue from a database, addresses states with both dog owners and veterinary professionals residing, requests poker player performance metrics, asks for details about authors affiliated with specific organizations and working in particular fields, inquires about states with specific types of residents, requests birth date of a specific player, focuses on specific data extraction, requests specific player details from a tennis database, involves joining multiple tables to extract information, focuses on querying player details in a tennis database, asks for author information based on affiliations and fields of study, aims to extract fix version for a particular issue, inquires about authors affiliated with a particular organization in a specific field, provides counts and lists based on specified conditions, focuses on retrieving specific information from a database, demonstrates the use of INTERSECT operator in SQL queries, requests specific details about a player in a sports database, identifies states with specific types of residents in a dog kennel database, inquires about the number of singles released in a specific year in a music database, asks for the count of singles released in a specific year from a music database, seeks information on states with specific types of residents in a dog kennel database, emphasizes the importance of accurate query reflection, focuses on retrieving specific data from a database, illustrates the importance of correct conditions in SQL queries, focuses on filtering data based on conditions, involves using SQL to retrieve data based on conditions, relates to analyzing data from databases
 \\
\end{longtable}
\end{center}
\twocolumn  

\end{document}